\documentclass[pdflatex,iicol]{sn-jnl}% APA Reference Style + Double Column

\usepackage{natbib}
\usepackage{graphicx}
\usepackage{multirow}
\usepackage{amsmath,amssymb,amsfonts,amsthm,mathrsfs}
\usepackage[title]{appendix}
\usepackage{xcolor}
\usepackage{textcomp}
\usepackage{manyfoot}
\usepackage{booktabs}
\usepackage{threeparttable}
\usepackage{listings}
\usepackage{tikz}
\usepackage{xspace}
\usepackage{wrapfig}
\usepackage{array}
\usepackage{makecell}
\usepackage{enumitem}
\usepackage{pifont}
\usepackage{hyperref}
\usepackage{url}
\usepackage{subcaption}
\usepackage{tabularx}

\usepackage{caption}

\usepackage{algorithm}%
\usepackage{algorithmicx}%
\usepackage{algpseudocode}%

\theoremstyle{thmstyleone}%
\theoremstyle{thmstyletwo}%

\theoremstyle{thmstylethree}%

\newcolumntype{Y}{>{\centering\arraybackslash}X}
\definecolor{cvprblue}{rgb}{0.21,0.49,0.74}
\usepackage{hhline}
\hypersetup{
    citecolor=cvprblue,linkcolor=red
}

\newcommand{\model}{\emph{{LENS}}\xspace}
\newcommand{\raa}[1]{\renewcommand{\arraystretch}{#1}}

\makeatletter
\DeclareRobustCommand\onedot{\futurelet\@let@token\@onedot}
\def\@onedot{\ifx\@let@token.\else.\null\fi\xspace}
\def\eg{\emph{e.g}\onedot}   
\def\ie{\emph{i.e}\onedot}   
\def\cf{\emph{c.f}\onedot}   

\def\vs{\emph{vs}\onedot}

\makeatother

\begin{document}

\title[LENS]{Robust Reasoning Segmentation for Open-World MLLMs without Catastrophic Forgetting}

\author{\fnm{Jiazhen} \sur{Liu}}\email{jliugj@connect.ust.hk}

\author*{\fnm{Long} \sur{Chen*}}\email{longchen@ust.hk}

\affil{\orgdiv{Department of Computer Science and Engineering}, \orgname{The Hong Kong University of Science and Technology}, \orgaddress{\street{Clear Water Bay}, \city{Kowloon}, \state{Hong Kong SAR}}}

\abstract{Ensuring trustworthiness and versatility in Open-World Visual Recognition is a core goal of Multimodal Large Language Models (MLLMs). While integrating segmentation is desirable, prevailing methods require invasive fine-tuning, which often leads to catastrophic forgetting of the model's general-purpose capabilities.  This trade-off between task-specific performance and general reliability fundamentally undermines the goal of building unified, trustworthy systems. We introduce \model (\textbf{L}everaging k\textbf{E}ypoi\textbf{N}ts for MLLMs' \textbf{S}egmentation), a novel plug-and-play paradigm designed to equip MLLMs with pixel-level ability without compromising their foundational integrity. \model attaches a lightweight head to a completely frozen MLLM, interpreting the intrinsic spatial cues embedded in attention maps. It extracts keypoints and translates them into prompt-compatible features for a mask decoder. Extensive experiments validate that \model achieves segmentation performance {competitive with} retraining-based methods. Crucially, it largely preserves the MLLM's open-world generalization, avoiding the degradation observed in fine-tuned counterparts. By decoupling segmentation from generation, \model offers a robust and efficient path toward truly multi-talented and reliable vision models.}

\keywords{Reasoning segmentation, generalization preservation, plug-and-play, MLLMs}

\maketitle

\section{Introduction}

Built on Large Language Models (LLMs), Multimodal LLMs (MLLMs) have demonstrated generalized visual understanding, most notably through their ability to ground language instructions in specific image regions~\citep{zhang2025mllms}. This property connects high-level semantics with {visual space}, paving the way to reformulate different vision tasks into a unified visual-instruction-controlling manner~\citep{wu2024dettoolchain, lai2024lisa, ma2024groma}. As this trend unfolds, MLLMs are expected to encompass a full spectrum of visual tasks, including recognition~\citep{llava}, detection~\citep{wu2024dettoolchain}, and even dense, pixel-level segmentation~\citep{lai2024lisa}. 

\begin{figure*}[t]
    \centering
    \includegraphics[width=0.9\linewidth]{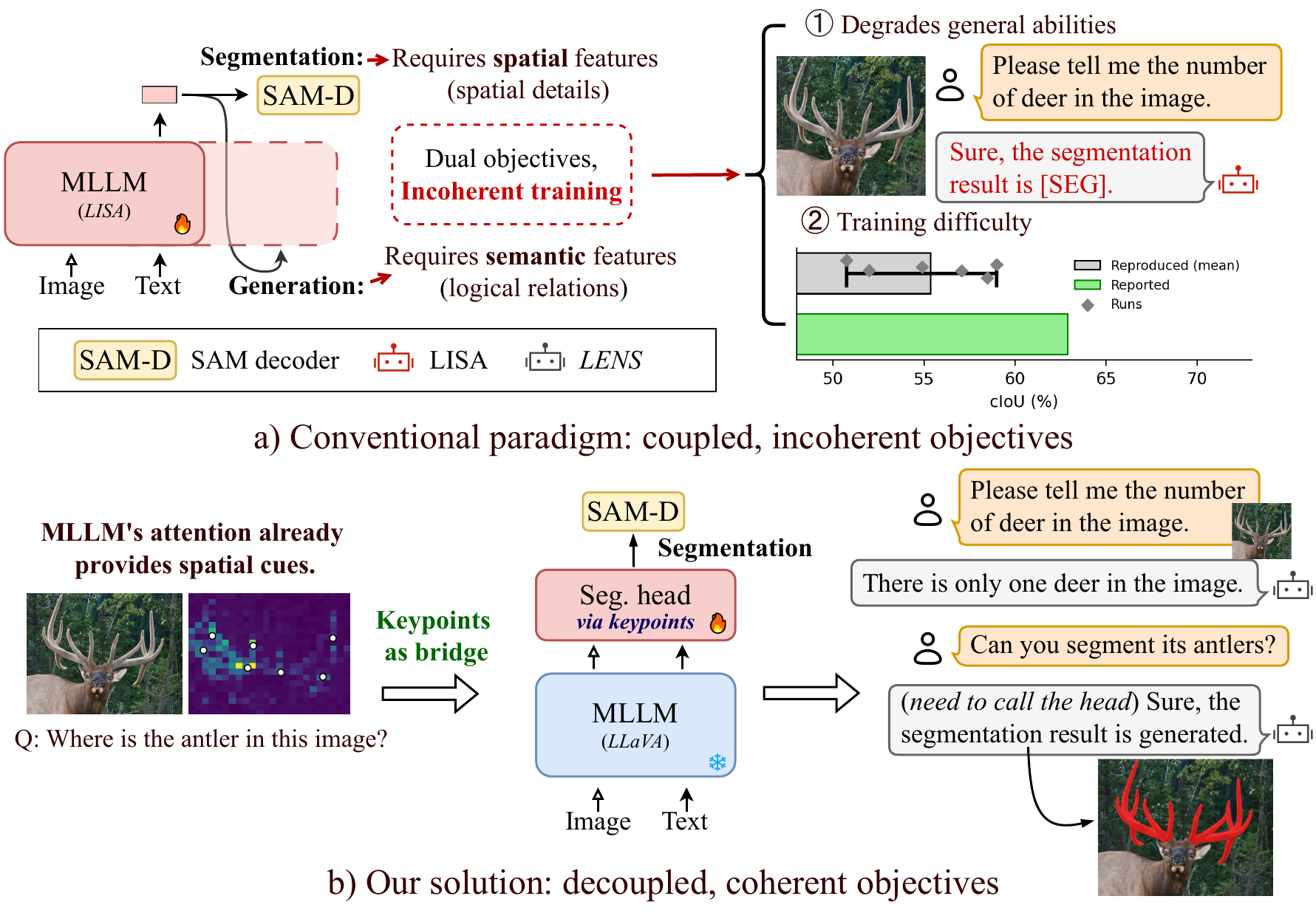}
    % \vspace{-2em}
\caption{{Conventional architecture for MLLM segmentation vs. \model.}  Conventional methods (\eg LISA~\citep{lai2024lisa}) fine-tune an MLLM for both generation and segmentation tasks, leading to conflicting objectives that undermine the model’s general capabilities and training stability. \model decouples these roles: a frozen MLLM is dedicated to reasoning, while a lightweight head is trained exclusively for segmentation. The head can be adaptively invoked by the model when needed, allowing the MLLM to serve as a unified vision model capable of handling diverse tasks}
    \label{fig:intro}
\end{figure*}

Yet, integrating segmentation capability presents a unique challenge, as its dense pixel-mask outputs cannot be natively expressed by the text-generative nature of LLMs, nor is there a large-scale segmentation corpus for autoregressive pre-training~\citep{lai2024lisa}. This skill must instead be transferred from a conventional, pre-trained segmentation model~\citep{lai2024lisa, read, wu2024see}. As illustrated in Fig.~\ref{fig:intro}a, prevailing approaches feed MLLM features into SAM's decoder~\citep{ravi2024sam}, which then maps them into masks. Notably, a significant mismatch exists: segmentation decoders are designed for low-level spatial cues (\eg, points or boxes), whereas MLLMs produce high-level, abstract semantic features~\citep{jiang2025devils}. To bridge this gap, existing solutions always involve extensively fine-tuning the MLLM with both segmentation and generation objectives, thereby training it to produce features compatible with the segmentation decoder~\citep{read}. Despite its straightforwardness, these approaches prove highly effective for instruction-controlled segmentation.

This effectiveness, however, comes at a cost. The dual-objective training introduces an inherent tension between model's capabilities: generative tasks thrive on abstract, sparse semantics, whereas segmentation requires direct, spatial features~\citep{liu2024insight}. Although large models can accommodate both, this compatibility is fragile and often leads to catastrophic forgetting of the model's original capabilities~\citep{wu2024see}. Take LISA~\citep{lai2024lisa} as an example, which is concurrently trained to generate a special \texttt{[SEG]} token and adapt its corresponding features to be compatible with the SAM-based decoder. Consequently, it frequently defaults to segmentation-focused responses like ``Sure, the segmentation result is \texttt{[SEG]}", even for a completely unrelated counting query (\cf, Fig.~\ref{fig:intro}a). This narrow focus reduces the MLLM to a single-task tool, causing its performance on the general-purpose benchmark like MMBench~\citep{liu2024mmbench} to plummet to near-zero. Such an outcome fundamentally contradicts the goal of building trustworthy and versatile vision models. 

Another drawback, as noted by prior studies~\citep{chen2024sam4mllm, zhu2025segagent}, is that combining segmentation and generation losses increases optimization complexity, which we also observe in our reproductions (\cf Fig.~\ref{fig:intro}a). Dual-objective models are highly sensitive to training configurations and require extensive hyperparameter tuning to achieve competitive results.

Motivated by these limitations, we argue that segmentation should be introduced as a \textbf{plug-and-play} capability, one that enhances the MLLM without compromising its robustness and open-world reasoning abilities. An intuitive strategy is to freeze the MLLM entirely and train an external head dedicated to converting its features for segmentation. However, this simple architectural change is insufficient. Frozen MLLMs provide only semantic features, having already discarded most of the fine-grained spatial details critical for segmentation~\citep{jiang2025devils}. This flaw requires more than a mere structural modification; it necessitates a paradigmatic shift in how MLLMs' features are leveraged.

Our approach sparks this shift through a crucial insight:  an MLLM's internal attention mechanisms already provide the spatial cues~\citep{jiang2025devils, wang2024mllm, zhang2025mllms}. As illustrated in Fig.~\ref{fig:intro}b, when an MLLM processes a query, a distinct attention pattern emerges over the image, with high-scoring regions corresponding to the object of interest (\eg, the antler). 
This allows us to repurpose the segmentation head for a more direct task: refining these attention-derived spatial cues into keypoint coordinates and using the MLLM's semantic judgment to assign corresponding descriptions (labels). These \textbf{keypoint--description} pairs act as direct prompts for the SAM decoder, effectively bridging the MLLM's internal representations with the segmentation model's input requirements. By leveraging the MLLM's native abilities for both localization via attention and verification via semantics, this process makes the head's training remarkably coherent.

We call this architecture \model (\textbf{L}everaging k\textbf{E}ypoi\textbf{N}ts for MLLMs' \textbf{S}egmentation), which equips MLLMs with segmentation capability while keeping the backbone entirely frozen. 
This design avoids degrading the MLLM’s general-purpose abilities and delivers substantial efficiency gains: Since the MLLM is used purely in inference mode, training costs are greatly reduced. Meanwhile, the segmentation head functions as a modular, plug-and-play tool that can be invoked on demand, enabling seamless integration into agent-based systems (\cf Fig.~\ref{fig:intro}b). 
Overall, our contributions can be summarized as follows:

\vspace{-0.5em}
\begin{enumerate}[leftmargin=*]
\itemsep-0.2em
\item We introduce \model, a novel segmentation architecture to operate on a completely frozen MLLM backbone. This decoupled paradigm prevents catastrophic forgetting, thereby ensuring the trustworthiness of the MLLM's general-purpose abilities while resolving a central flaw in prior fine-tuning methods.
\item We demonstrate how spatial cues from an MLLM’s internal attention can be refined into SAM-compatible prompts, with keypoints serving as the bridge between high-level reasoning and pixel-level segmentation.
\item \model achieves {performance on par with state-of-the-art fine-tuning-based methods under the same lightweight training setting} on multiple segmentation benchmarks while notably reducing training costs, as the core MLLM is utilized purely for inference. Its efficiency and plug-and-play design offer a practical and scalable solution for unified vision models.

\end{enumerate}

\section{Related Work}

\noindent\textbf{Multimodal Large Language Models (MLLMs).} The advent of MLLMs represents a paradigm shift in computer vision, driven by the powerful reasoning capabilities inherited from their underlying LLMs~\citep{kaplan2020scaling, openai2024hello, gemini}. Architectures like LLaVA~\citep{llava, liu2024improved}, InstructBLIP~\citep{dai2023instructblip}, and Qwen-VL~\citep{bai2023qwenvl} typically connect a visual encoder to a pre-trained LLM core via lightweight, parameter-efficient modules, and can be further adapted to domain-specific applications~\citep{yao2025llava}. This architectural integration enables the generation of text grounded in visual input, establishing a robust and sophisticated alignment between language and vision. 

There are two primary ways to leverage this intrinsic vision-language spatial association: either the model directly articulates its understanding through generated text~\citep{bai2025qwen2, peng2023kosmos}, or its internal mechanisms, such as attention, can be decoded to reveal its spatial cues~\citep{zhang2025mllms, wang2024mllm}.

\noindent\underline{\emph{Spatial Cues in Attention.}} Recent investigations have consistently shown that the attention mechanisms within MLLMs serve as a natural bridge between textual tokens and their corresponding image regions~\citep{zhang2025mllms, wang2024mllm, yang2025look, yu2024attention, kang2025see}. When conditioned on paired image-text input, attention maps highlight the regions most relevant to the textual description, effectively providing coarse spatial cues of the target~\citep{wang2024mllm}. Crucially, this is not an idiosyncratic feature of any single architecture but a universal, emergent property observed across a diverse range of models~\citep{wang2024mllm, zhang2025mllms, yu2024attention, kang2025see}. This phenomenon arises organically from the model's objective to generate text that is contextually grounded in the visual input; to accurately describe an object, the model must first  ``look'' at it. Consequently, attention maps offer a robust and direct source of spatial information, making them an ideal foundation for dense prediction tasks like segmentation. Moreover, leveraging these explicit attention patterns inherently enhances the interpretability of the system, offering a transparent glimpse into the model's decision-making process, a key factor for trustworthy AI.

\noindent\textbf{Segmentation Models.} Early image segmentation paradigms, such as semantic and panoptic segmentation~\citep{badrinarayanan2017segnet, long2015fully, ronneberger2015u}, were predominantly closed-set, operating on a fixed vocabulary of object categories. A recent shift towards open-world perception and open-vocabulary segmentation has been driven by promptable models that accept diverse control signals~\citep{ravi2024sam, kirillov2023segment, liu2023referring, wu2024toward, ren2024grounded, zou2023generalized, liu2023gres, zou2023segment}. These range from low-level spatial prompts (\eg, points, boxes) in models like SAM~\citep{kirillov2023segment} to explicit textual phrases in Referring Expression Segmentation (RES)~\citep{res, liu2023referring, wu2024toward, ren2024grounded, liang2023open, zou2023generalized}. Despite their flexibility, these methods are fundamentally limited by their dependence on direct, literal prompts. They lack the higher-level reasoning ability needed to ground complex, inferential semantics in pixel space, which motivates the development of dedicated reasoning-based segmentation models.

\noindent\underline{\emph{Reasoning Segmentation Models.}} As an advanced form of RES, reasoning segmentation targets objects that are only implicitly referenced and must be inferred from descriptive cues (\eg, segment “the organ used for defense” instead of just “antler”). The inherent demand for strong comprehension and reasoning has naturally positioned MLLMs as the foundational backbone for this task~\citep{lai2024lisa, rasheed2024glamm, ren2024pixellm, wu2024see, xia2024gsva, read}. LISA~\citep{lai2024lisa} pioneered this task by training an MLLM to emit a special token whose feature is then fed into a SAM-like decoder; the model is jointly optimized on large-scale mixtures of instruction-following and segmentation data to transfer the MLLM's reasoning ability to the segmentation domain. Building on this paradigm, SESAME~\citep{wu2024see} introduces negative examples to enable refusal of non-segmentable queries, while READ~\citep{read} analyzes the underlying mechanism and proposes similarity-based objectives to further refine performance. Although viable, these methods all rely on heavy joint training. Even with optimizations like LoRA~\citep{hu2022lora, mao2025survey}, tightly coupling the objectives for generation (semantics) and segmentation (spatial) creates a trade-off. This often leads to the MLLM becoming over-specialized, inducing catastrophic forgetting of its general-purpose abilities. This degradation severely undermines the trustworthiness and reliability of the model when deployed in complex, open-world scenarios.

\section{Proposed \model}

In this section, we present \model, a novel architecture that equips a frozen MLLM with segmentation in a plug-and-play manner. As illustrated in Fig.~\ref{fig:method}a, \model consists of three stages: a lightweight head (\S\ref{sec:head}), a keypoints extraction and description module (\S\ref{sec:keypoint}), and a mask decoder (\S\ref{sec:decoder}). 
The central innovation of this design is the use of keypoints from the MLLM’s internal attention maps as a bridge that intrinsically unifies the stages. We next detail each stage, followed by the training objectives and configurations (\S\ref{sec:training}). Key notations used throughout this section are summarized in Table~\ref{tab:notations}.

% --- 插入表格代码 ---
% --- 插入符号定义表 Start ---
\begin{table}[t]
    \centering
    \small
    \caption{{Notations.} Summary of key symbols used in the \model\ architecture (\cf Fig.~\ref{fig:method})}
    \label{tab:notations}
    % 使用 tabularx 将表格宽度锁定为 \linewidth (即当前栏宽)
    % 'X' 列会自动计算宽度并换行
    \begin{tabularx}{\linewidth}{l X} 
        \toprule
        \textbf{Symbol} & \textbf{Description} \\
        \midrule
        $I, T$ & Input image and instruction text \\
        $F_i, F_t$ & Intermediate MLLM features for image and text \\
        $H_{\text{in}}^1$ & Input features to the head (concatenation of $F_i$ and $F_t$) \\
        $A_c^1$ & Text-to-image attention from the start-of-answer token (grounding map) \\
        $H_{\text{out}}^1, H_{\text{out}}^2$ & Output features of Head Layer 1 and Layer 2 \\
        $f_s^2$ & Feature of the start-of-answer token (used as global query) \\
        $P$ & Set of sub-pixel refined keypoint coordinates \\
        $d_i$ & Semantic description for keypoint $i$ \\
        $D$ & Final set of descriptions served as label embeddings \\
        $P^{\text{pos}}$ & Position embeddings encoded from $P$ \\
        $\hat{M}$ & Final predicted segmentation mask \\
        \bottomrule
    \end{tabularx}
\end{table}
\subsection{Segmentation Head}
\label{sec:head}

The segmentation head receives semantic features from the MLLM, (i) refines the attention dependencies to increase target-region keypoints, and (ii) provides a decision on whether the attention-highlighted regions should be identified as segmentation targets. 

\noindent\textbf{Architecturally}, the head is a two-layer transformer, mirroring a single MLLM layer for consistency. 
Its dual roles impose two requirements on input features: (i) strong cross-modal attention\footnote{Attention from text to image tokens.}, and (ii) sufficient semantics to identify grounded targets. 
As shown in prior work~\citep{zhang2025mllms, jiang2025devils}, shallow layers are deficient in semantics, while deep layers exhibit diminished cross-modal attention. 
Thus, we adopt intermediate features (\eg, the 14th layer in LLaVA-1.5-7B), which best balance these properties.

\noindent{\textbf{Scope of the Frozen Setting.}
\model targets a \emph{frozen-but-accessible} setting: the MLLM is kept entirely frozen and none of its parameters is updated, but its intermediate attention maps and features are assumed to be accessible for deriving keypoints. \model is therefore not applicable to strictly black-box, API-only models that expose text outputs alone. This scope does not affect our main claim, which is to preserve the MLLM backbone and avoid segmentation-specific fine-tuning of the language model (see also \S\ref{sec:limitations}).}

\begin{figure*}[t]
    \centering
    \includegraphics[width=\textwidth]{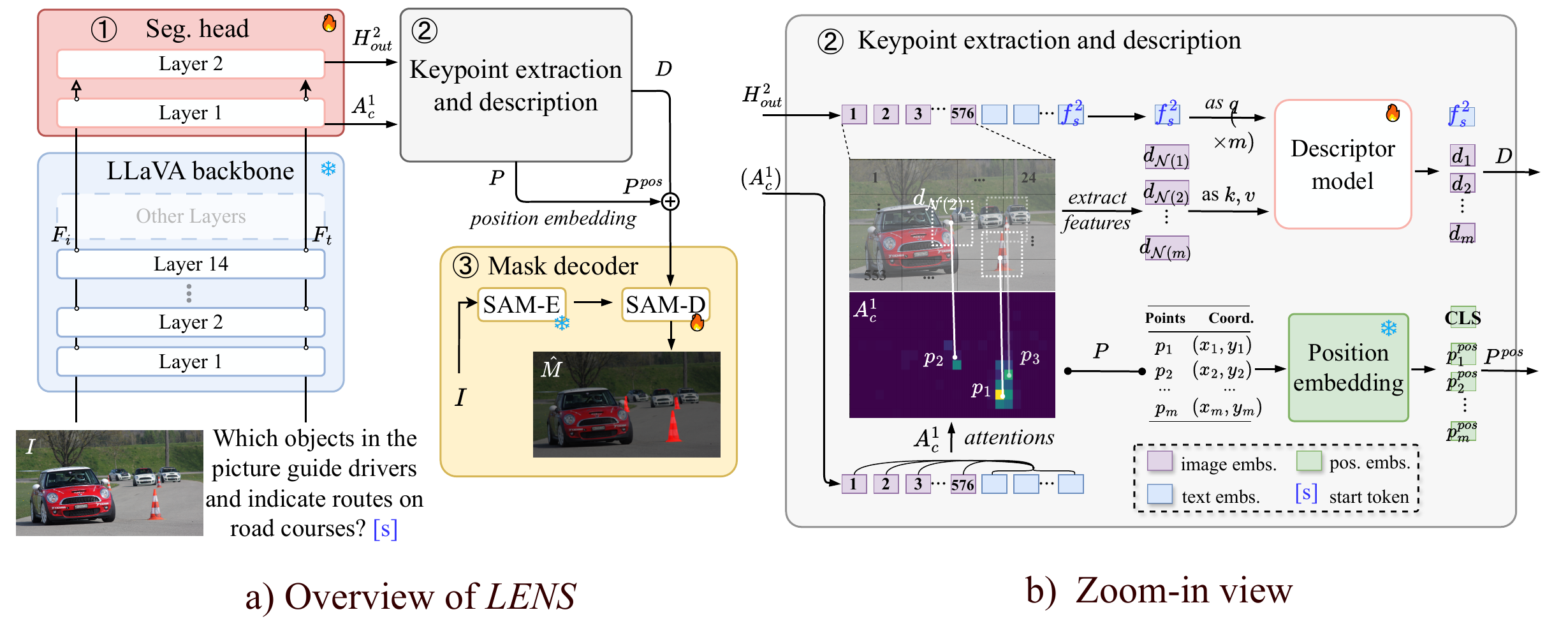}
    % \vspace{-2em}
    \caption{{\model\ architecture.} \textbf{(a)} Overall architecture built on LLaVA-1.5-7B~\citep{liu2024improved}, consisting of three stages: \ding{172} a segmentation head that refines attention and semantic features, \ding{173} keypoint extraction and description, and \ding{174} a mask decoder that takes fused keypoint descriptions and coordinates to predict the final mask. Trainable and frozen modules are indicated, and the MLLM runs only in inference mode. \textbf{(b)} Zoom-in view of the attention and keypoint components}
    \label{fig:method}
\end{figure*}

Given an input image $I$ and instruction $T$, we denote their intermediate features as $F_i \in \mathbb{R}^{L_i \times d}$ for the image (with $L_i=576$ in LLaVA-1.5-7B) and $F_t \in \mathbb{R}^{L_t \times d}$ for the text. These are concatenated into $H^{1}_{\text{in}} = [F_i; F_t] \in \mathbb{R}^{L \times d}$ which serves as the input to the head. For simplicity of exposition, we assume a fixed ordering where text features are always appended after the image features.

\noindent\textbf{Layer 1: Attention Refinement}. While MLLM attention maps can localize grounded targets, they are not tailored for segmentation. As observed by~\citet{darcet2024vision}, they often highlight contextual regions useful for text generation but irrelevant to segmentation. Therefore, the first layer is to re-calculate and refine these maps, explicitly training them to suppress extraneous activations and selectively highlight only the regions corresponding to the intended target. This explicit refinement not only aids segmentation but also enhances the interpretability of the model's focus, providing transparent evidence of its reasoning process.

To achieve this, the layer first computes a full attention map $A^1$ over all input tokens. We then take the attention from the \emph{start}-of-answer token (\ie, the final token of the input query) to the image tokens, yielding the text-to-image grounding map $A_c^1$ (Fig.~\ref{fig:method}b). The computation is as follows:
\begin{equation}
\left\{
\begin{aligned}
    A^1 &= \text{Softmax}\left(\frac{QK^\top}{\sqrt{d}} + M_a\right), \\
    A_c^1 &= A^1[L_i+L_t,\; 1:L_i].
\end{aligned}
\right.
\label{eq1}
\end{equation}
where $Q$ and $K$ are linear projections of input features $H_{\text{in}}^1$, and $M_a$ is the causal attention mask. $A_c^1$ is explicitly optimized through training, and $A^1$ is used to produce the output features $H_{\text{out}}^1 \in \mathbb{R}^{L \times d}$.

\noindent\textbf{Layer 2: Feature Enhancement.} Since the attention in Layer 1 is explicitly optimized, the resulting representation $H_{\text{out}}^1$ may carry semantic bias. Layer~2 aims to mitigate this bias and enhance the discriminative semantics of $H_{\text{out}}^1$, thereby producing the output $H_{\text{out}}^2$. We expect the feature of the \emph{start}-of-answer token\footnote{Derived from the question’s final token \texttt{[s]} in Fig.~\ref{fig:method}b.} $f_s^2$ to align with the image features in $H_{\text{out}}^2$, enabling it to serve as a semantic query for identifying the image regions that correspond to segmentation targets.

Overall, the process of the segmentation head can be summarized formally as follows:
\begin{equation}
\left\{
\begin{aligned}
    A^1,\, H_{\text{out}}^1 &\gets \text{Layer}_1([F_i;F_t]), \\
    A_c^1 &\gets A^1[L_i+L_t,\; 1:L_i], \\
    H_{\text{out}}^2 &\gets \text{Layer}_2(H_{\text{out}}^1).
\end{aligned}
\right.
\end{equation}

\subsection{Keypoint Extraction and Description}
\label{sec:keypoint}

The second stage extracts points from the high-value regions of $A_c^1$, which serve as indicators for segmentation.
We define these as \emph{keypoint} regions. Each keypoint is then \emph{described} as positive if the semantics of its image feature in $H_{\text{out}}^2$ match $f_s^2$, and negative otherwise. The resulting positions and descriptions together form the prompts, which serve as the structured input to the SAM decoder (\cf Fig.~\ref{fig:method}b).

\noindent\textbf{Keypoint Extraction.}
The attention map $A_c^1$ is reshaped into a 2D heatmap { and bilinearly resized to a finer grid (we use the same $128 \times 128$ resolution at which $\mathcal{L}_{\text{attn}}$ is applied; \cf \S\ref{sec:training})}, from which keypoints are extracted via Non-Maximum Suppression (NMS). Local maxima are selected as candidate positions, and up to $m$ keypoints are retained.

Since $A_c^1$ is defined at the patch level, the heatmap resolution is low and coordinates are confined to grid positions, which is suboptimal for pixel-level segmentation. To mitigate this, we apply a \emph{sub-pixel} refinement that shifts grid-aligned coordinates toward the underlying peak locations. The refined set is denoted $P=\{p_i\}_{i=1}^{m}$, where each $p_i=(x_i,y_i)$. These keypoints are then encoded into position embeddings $P^{pos}$ compatible with the SAM decoder. The implementation of this encoding is deferred to \S\ref{sec:decoder}.

\begin{figure*}[t]
    \centering
    \includegraphics[width=\linewidth]{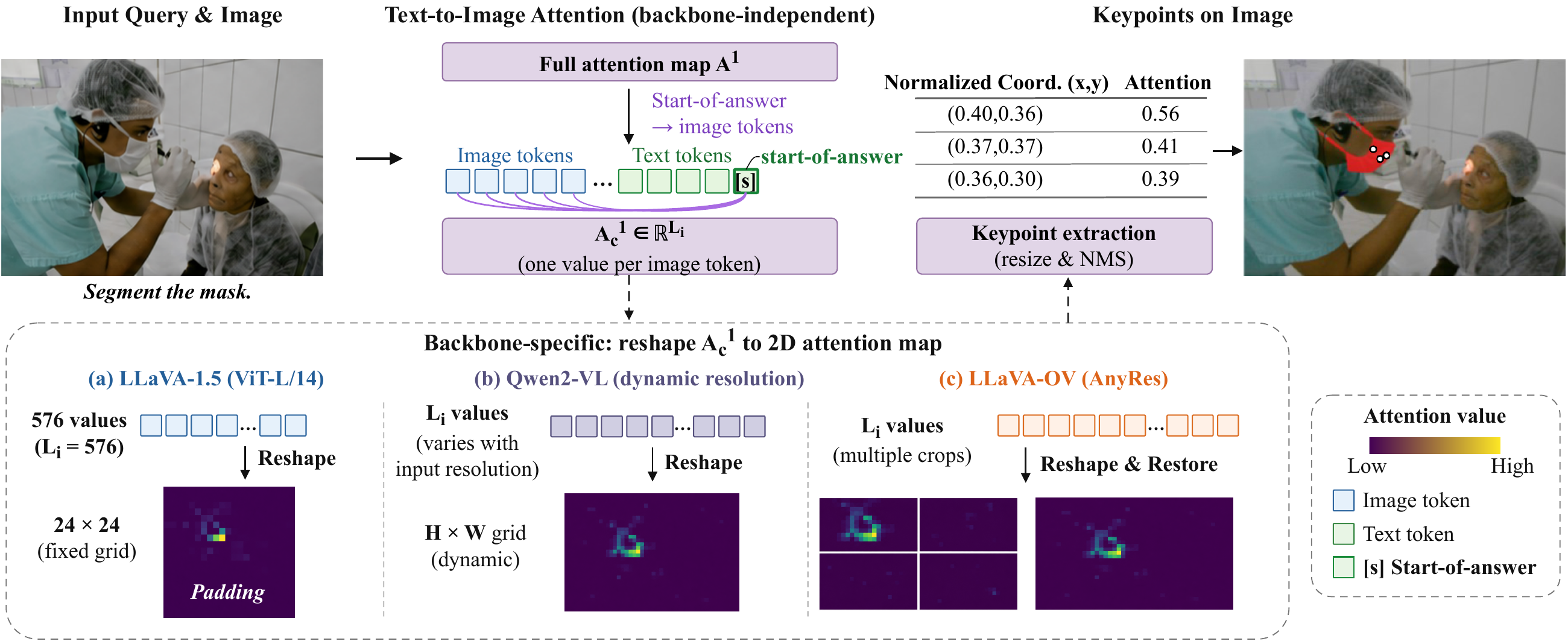}
\caption{{Tracing attention back to image patches across different MLLM backbones.}
The attention map $A_c^1$ is computed identically across backbones, while its spatial reshaping follows the corresponding visual tokenization strategy: a fixed $24\times24$ grid for LLaVA-1.5, a variable-size grid for Qwen2-VL, and an AnyRes crop arrangement for LLaVA-OV. The recovered 2D attention map is then processed by the same keypoint extraction pipeline}
    \label{fig:attn_show}
\end{figure*}

\noindent{\textbf{Attention-to-Patch Correspondence.}
A natural question is how attention over visual tokens can be traced back to image coordinates across MLLMs with different visual tokenization strategies. As defined in Eq.~\ref{eq1}, given $L_i$ visual tokens and $L_t$ text tokens, we compute the attention map $A^1$ and take the attention from the start-of-answer token to all visual tokens, obtaining $A_c^1 \in \mathbb{R}^{L_i}$. This attention computation is identical across the evaluated MLLM backbones and does not depend on how the image is resized, divided, or tokenized. Mapping these $L_i$ attention values back to image space simply follows the spatial organization of the visual tokens defined by the corresponding backbone's visual tokenization process.

Taking LLaVA-1.5 as an example, the CLIP ViT-L/14 encoder maps the padded $336 \times 336$ input to a fixed $24 \times 24$ grid of 576 visual tokens. Since its projector operates \emph{token-wise} without mixing spatial positions, the length-576 attention map $A_c^1$ can be directly reshaped into a $24 \times 24$ heatmap. For Qwen2-VL, the number and spatial arrangement of visual tokens vary with the input resolution, so we retain the corresponding visual-token grid produced during preprocessing and reshape the attention values according to this grid. For LLaVA-OV, the image is processed through its AnyRes strategy, and we similarly retain the organization of visual tokens across image crops and restore the attention values according to the corresponding crop arrangement. Therefore, although different MLLMs may organize visual tokens differently, the computation of $A_c^1$ remains unchanged, and tracing it back to image space only requires following the spatial arrangement already defined by each backbone, as illustrated in Fig.~\ref{fig:attn_show}. Once the 2D attention map is obtained, the same NMS and sub-pixel refinement are applied to extract keypoints.
}

\noindent\textbf{Sub-pixel Coordinate Refinement.}
While the attention map $A_c^1$ provides coarse localization, its resolution is limited by the patch grid (\eg, $24 \times 24$). To achieve pixel-level precision required for segmentation, we refine the extracted integer keypoints $\mathbf{p}^{\text{int}}$ via a second-order Taylor expansion.
Let $H(\mathbf{x})$ denote the continuous heatmap function approximated by the discrete attention map. For a candidate keypoint $\mathbf{p}_k$, we define the local objective using a Taylor series expansion around $\mathbf{p}_k$:
\begin{equation}
    H(\mathbf{p}_k + \boldsymbol{\delta}) \approx H(\mathbf{p}_k) + \mathbf{g}^\top \boldsymbol{\delta} + \frac{1}{2} \boldsymbol{\delta}^\top \mathbf{H} \boldsymbol{\delta},
\end{equation}
where $\mathbf{g}$ and $\mathbf{H}$ are the gradient vector and Hessian matrix, respectively, estimated via discrete finite differences from the $3\times3$ neighborhood \citep{lowe2004distinctive}. The optimal sub-pixel offset $\boldsymbol{\delta}^*$ is obtained by maximizing this local approximation, leading to a closed-form Newton update:
\begin{equation}
    \boldsymbol{\delta}^* = -(\mathbf{H} + \varepsilon \mathbf{I})^{-1} \mathbf{g},
\end{equation}
where $\varepsilon$ is a regularization term for numerical stability. The final coordinates are updated as $\mathbf{p}^{\text{sub}}_k = \mathbf{p}^{\text{int}}_k + \boldsymbol{\delta}^*$. This refinement step is differentiable and computationally negligible, yet it significantly mitigates the quantization error inherent in patch-based attention.

\noindent\textbf{Keypoint Description.}
To determine whether each keypoint corresponds to a positive or negative region, we extract its associated semantic features. Specifically, at each coordinate we \emph{sample} the image feature from $H_{\text{out}}^2$ via interpolation, and further sample from a $3 \times 3$ neighborhood to enrich the semantic representation. This yields a local feature set $d_{\mathcal{N}(i)}$ for each keypoint $p_i$. 

\begin{figure}[t]
    \centering
    % 请确保文件名对应你原本补充材料里的图片路径
    \includegraphics[width=\linewidth]{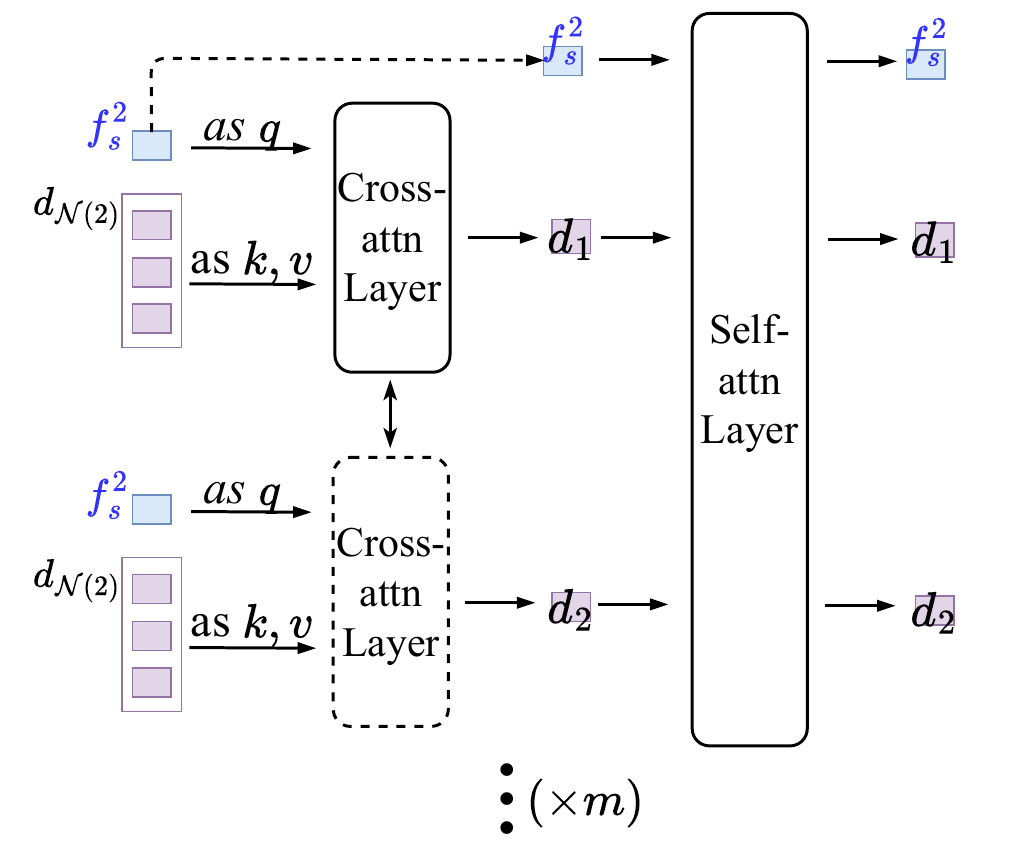} 
    \caption{{Detailed architecture of the Descriptor Model.} 
    Illustrating the interaction between the global semantic query ($f_s^2$) and local neighborhood features ($d_{\mathcal{N}(i)}$). The cross-attention mechanism serves as a verifier to determine if a keypoint corresponds to the target object, filtering out background noise effectively}
    \label{fig:descriptor_details}
\end{figure}

We then leverage the global \emph{start}-of-answer token feature $f_s^2$ as a query to determine whether the region of $p_i$ should be segmented.
The neighborhood features $d_{\mathcal{N}(i)}$ serve as keys and values in a descriptor model (\cf Fig.~\ref{fig:descriptor_details}), where \emph{cross-attention} is performed to produce discriminative descriptions $\{d_i\}_{i=1}^m$. 
Through the interaction between $f_s^2$ and neighborhood features, these descriptions are expected to acquire the discriminative capacity needed for positive/negative interpretation.

\noindent\textbf{Global Description.} 
While each keypoint yields a local description, these remain independent and may contain redundancy or spatial overlap. 
To promote coherence among them, we further introduce $f_s^2$ as a global semantic descriptor within the descriptor model. 
Through a subsequent \emph{self-attention} operation, $f_s^2$ interacts with all local descriptions $\{d_i\}_{i=1}^m$, enabling global context to regularize redundant or spatially overlapping instances while simultaneously consolidating information back into $f_s^2$. 
The final description set is defined as $D$.

\noindent\textbf{The Descriptor Model.}  
To determine the semantic relevance of each keypoint, we design a dedicated descriptor module (visualized in Fig.~\ref{fig:descriptor_details}).
% 这里可以插入原补充材料 Fig S1 作为正文的一个子图或单独的图
Unlike standard designs, our descriptor leverages both local context and global guidance:
\begin{itemize}
    \item \textbf{Local-Global Interaction:} For each keypoint $i$, we sample a set of local features $d_{\mathcal{N}(i)}$ from the feature map $H^2_{\text{out}}$. Using the global start-of-answer token $f_s^2$ as the \textit{query} and $d_{\mathcal{N}(i)}$ as \textit{keys/values}, we perform cross-attention to strictly condition the local description on the global semantic intent.
    \item \textbf{Global Contextualization:} The resulting descriptors $\{d_i\}_{i=1}^m$ are concatenated with $f_s^2$ and processed by a self-attention layer. This allows the model to suppress redundant keypoints and reason about spatial relationships before decoding the final mask.
\end{itemize}

The process of keypoint extraction and description can be summarized as:
\begin{equation}
\left\{ % 左边的大括号
\begin{aligned}
    P &\leftarrow \text{Sub-pixel}(\text{NMS}(A_c^1)), \\
    \{d_{\mathcal{N}(i)}\}_{i=1}^m &\leftarrow \text{Sample}(\mathbf{F}_i^2,\, P), \\
    \{d_i\}_{i=1}^m &\leftarrow \text{Cross-attn}(f_s^2,\, \{d_{\mathcal{N}(i)}\}_{i=1}^m), \\
    D &\leftarrow \text{Self-attn}(\{f_s^2\} \cup \{d_i\}_{i=1}^{m}).
\end{aligned}
\right. % 右边隐形闭合（必须有这个点）
\end{equation}

where both $\mathbf{F}_i^2$ (image tokens) and $f_s^2$ (start token) are taken from $\mathbf{H}_{\text{out}}^2$.

\subsection{Mask Decoder}
\label{sec:decoder}

At this stage, we have the keypoint set $P \in \mathbb{R}^{m \times 2}$ and the description set 
$D \in \mathbb{R}^{(m+1) \times d_s}$, where $d_s$ matches the embedding dimension of the SAM decoder.
The keypoints naturally match the point-based prompts of SAM, and the descriptors play the role of label embeddings. 
This structural alignment allows our outputs to be seamlessly integrated into the SAM decoder.

\noindent\textbf{Position Embedding.} 
We adopt SAM’s point \emph{position encoder} to transform the keypoints $P$ into embeddings $P^{\text{pos}}=\{p_i^{\text{pos}}\}_{i=1}^m$. 
Since the global descriptor $f_s^2$ lacks a spatial position, we introduce a learnable \texttt{[CLS]} embedding as its positional counterpart. 
This yields both the position embeddings $P^{\text{pos}}$ and the label embeddings $D$ required by the SAM decoder. 

The summed embeddings are fed into the decoder to generate the final mask $\hat{M}$:
\begin{equation}
\left\{
\begin{aligned}
    P^{\text{pos}} &\gets \{\text{CLS}\} \cup \text{PosEnc}(P), \\
    \hat{M} &\gets \text{Decoder}\!\left(D \oplus P^{\text{pos}},\, F_\text{img}^{\text{SAM}}\right).
\end{aligned}
\right.
\end{equation}
where $\oplus$ denotes element-wise addition and $F_{\text{img}}^{\text{SAM}}$ are the image features from the SAM encoder.

\subsection{Integration into Multimodal Agents}
\label{sec:agent_integration}

A distinctive advantage of \model over retraining-based methods (\eg, LISA) is its modularity and safety. Since the MLLM remains frozen, its general-purpose knowledge and guardrails are preserved intact, ensuring trustworthy behavior during interactions. \model can thus be integrated into agent frameworks as a reliable invocation tool rather than a monolithic black box.
Algorithm~\ref{alg:agent_seg} details the inference pipeline. The MLLM acts as a router, classifying the user intent into \textit{Dialogue}, \textit{Segmentation}, or \textit{Follow-up}. 

\begin{itemize}
  \item \textbf{Dialogue:} The instruction is a general conversational query unrelated to segmentation.  
  In this case, the agent directly performs autoregressive generation conditioned on the text and image, and outputs a natural-language answer.
  
  \item \textbf{Segmentation:} The instruction explicitly asks to segment certain objects or regions in the image.  
  The agent extracts an intermediate embedding from its internal representation and passes it to \model's head $\mathcal{H}$, which decodes a segmentation mask.  
  The segmentation result is visualized and stored in the memory $\mathcal{M}$ for potential future reference, and the system returns a fixed textual response together with the visualized result.

  \item \textbf{Follow-up:} The instruction refers to the previously segmented content (\eg, asking about the segmented object).  
  The original image and its segmentation result are concatenated and passed back to the agent $\mathcal{A}$, which then answers the follow-up question based on both.
\end{itemize}
This design enables seamless switching between general dialogue and vision-centric segmentation tasks, while maintaining conversational context through memory.

\begin{algorithm}[h]
\caption{Inference Pipeline for Agent-guided Segmentation}
\label{alg:agent_seg}
\begin{algorithmic}[1] % [1] 表示显示行号

% \KwIn 对应 \Require
\Require Instruction $u$, image $I$; Agent model $\mathcal{A}$; \model's head $\mathcal{H}$; memory $\mathcal{M}$

% 普通行需要加 \State
\State $\text{intent} \gets \mathcal{A}.\text{Route}(u,I)$ \Comment{$\in\{\text{dialogue},\text{seg},\text{followup}\}$}

% \If{条件}{内容} 变为 \If{条件} ... \EndIf
\If{$\text{intent}=\text{seg}$} \Comment{The user instruction contains a segmentation intent}
    \State $e \gets \mathcal{A}.\text{Embed}(u,I)$
    \State $\text{mask} \gets \mathcal{H}.\text{Decode}(e,I)$
    \State $\mathcal{M}.\text{last} \gets (I,\text{mask})$
    \State \textbf{return} ``Sure, the segmentation result is generated.'', $\text{Overlay}(I,\text{mask})$
% \ElseIf{条件} 变为 \ElsIf{条件} (注意拼写是 ElsIf)
\ElsIf{$\text{intent}=\text{followup}$}
    \State $(I_0,\text{mask}_0) \gets \mathcal{M}.\text{last} \ \text{or}\ (I,\varnothing)$
    \If{$\text{mask}_0=\varnothing$}
        \State \textbf{return} \textsc{ThisAlgorithm}($u,I$)
    \EndIf
    \State $C \gets \text{Concat}(I_0, \text{Overlay}(I_0,\text{mask}_0))$
    \State \textbf{return} $\mathcal{A}.\text{Generate}(u, C)$
\Else
    \State \textbf{return} $\mathcal{A}.\text{Generate}(u, I)$
\EndIf

\end{algorithmic}
\end{algorithm}

\subsection{Training and Usage}
\label{sec:training}
\noindent\textbf{Training.}
Our model is trained end-to-end with a composite loss function consisting of two components: 
an attention loss $\mathcal{L}_{\text{attn}}$ and a segmentation loss $\mathcal{L}_{\text{seg}}$.

\noindent\underline{\emph{Attention Loss.}}
$\mathcal{L}_{\text{attn}}$ provides direct supervision for the cross-modal attention map $A_c^1 \in [0,1]^{h \times w}$. 
Given the ground-truth binary mask $M \in \{0,1\}^{h \times w}$, 
we use the binary cross-entropy loss to enforce alignment between $A_c^1$ and $M$:
\begin{equation}
\begin{split} % 使用 split 环境，编号居中
    \mathcal{L}_{\text{attn}} = - \frac{1}{hw} \sum_{i=1}^{h} \sum_{j=1}^{w} \Big[ & M_{i,j} \log A^1_{c,i,j} \\
    & + (1 - M_{i,j}) \log \big(1 - A^1_{c,i,j}\big) \Big].
\end{split}
\end{equation}
{In practice, the text-to-image attention vector $A_c^1$ is reshaped to a $24 \times 24$ spatial grid and bilinearly resized to $h \times w = 128 \times 128$, and the ground-truth mask $M$ is resized to the same resolution, where the attention loss is applied.}

\noindent\underline{\emph{Segmentation Loss.}}
For the segmentation loss $\mathcal{L}_{\text{seg}}$, we follow the practice of LISA~\citep{lai2024lisa} 
and adopt a combination of Dice loss and BCE loss applied to the final predicted mask $\hat{M} \in [0,1]^{h \times w}$. 
It's the weighted sum of the Dice and BCE losses:
\begin{equation}
    \mathcal{L}_{\text{seg}}
    = \lambda_{\text{dice}} \mathcal{L}_{\text{dice}}
    + \lambda_{\text{bce}} \mathcal{L}_{\text{bce}}.
\end{equation}
\underline{\emph{Overall Objective.}}
The overall training objective combines the two losses:
\begin{equation}
    \mathcal{L} 
    = \mathcal{L}_{\text{seg}}(\hat{M}, M) 
    + \mathcal{L}_{\text{attn}}(A_c^1, M).
\end{equation}
\noindent\textbf{Usage.}
Unlike token-based triggering mechanisms, \model relies on the MLLM to determine through question answering whether segmentation should be activated. The routing can be implemented using tool frameworks~\citep{langchain_official_website, wang2024survey} or thinking-based control~\citep{liu2025visionreasoner}. {A quantitative evaluation of this routing mechanism, including its robustness to prompt phrasing and mixed-intent queries, is presented in \S\ref{sec:robustness}.}

\section{Experiments}

\subsection{Experimental Setup}

\textbf{Implementation Details.}  
For a fair comparison with prior works, we adopt the widely used LLaVA-1.5-7B~\citep{liu2024improved} as the default backbone for the main experiments, with SAM instantiated by ViT-H. In addition, we report results with alternative MLLM backbones, including LLaVA-OV~\citep{li2024llava} and Qwen2-VL~\citep{wang2024qwen2vl}, as well as different SAM variants, including ViT-B and ViT-L, to further evaluate the generalization ability of \model across backbone and decoder choices. {Porting \model to a new backbone leaves its core design unchanged---the keypoint pipeline, descriptor, training objectives, and keypoint$\rightarrow$SAM-prompt interface are all kept identical; only two backbone-dependent dimensions are re-fit: the two-layer head is re-instantiated to mirror one transformer layer of the new backbone, and the intermediate feature layer is re-selected around its middle (the 14\textsuperscript{th} layer is specific to LLaVA-1.5-7B).} We used the 14\textsuperscript{th} layer as the intermediate representation, set $m=16$, and adopted a neighborhood size of $3 \times 3$. The optimizer was AdamW with a learning rate of $5\times10^{-5}${, betas $(0.9, 0.95)$, and zero weight decay; the learning rate was kept constant throughout training, with no warmup or decay. Training ran for 10 epochs of 100 steps each with an effective batch size of 256 ($8$ GPUs $\times$ $32$), \ie, 256k training samples in total}. The loss weights for Dice and BCE in $\mathcal{L}_{\text{seg}}$ were set to 4 and 2, respectively. {For keypoint sampling, candidate keypoints were extracted as local maxima of the resized attention map via NMS, retaining up to $m$ peaks; each keypoint is described as positive if its descriptor matches the start-of-answer query and negative otherwise (\cf \S\ref{sec:keypoint}).} Unless otherwise specified, other training settings followed LISA~\citep{lai2024lisa}. 

\noindent\textbf{Training Datasets.}  
\label{trainingd}
Following the dataset organization in LISA, we considered three categories: (1) semantic segmentation datasets including ADE20K~\citep{zhou2019semantic}, COCO-Stuff~\citep{caesar2018coco}, PACO-LVIS~\citep{ramanathan2023paco}, PASCAL-Part~\citep{chen2014detect}, and Mapillary Vistas~\citep{neuhold2017mapillary}; (2) referring segmentation datasets including RefCLEF, RefCOCO, RefCOCO+~\citep{kazemzadeh2014referitgame}, and RefCOCOg~\citep{he2017mask}; and (3) reasoning segmentation dataset ReasonSeg~\citep{lai2024lisa}. {During training, the three categories are sampled with ratios $9{:}1{:}1$ (semantic : referring : reasoning); within each category, the constituent datasets are sampled uniformly.} Note that \model was trained only with segmentation objectives and preserves general abilities without extra VQA data.

\begin{figure*}[t]
    \centering
    \includegraphics[width=.88\textwidth]{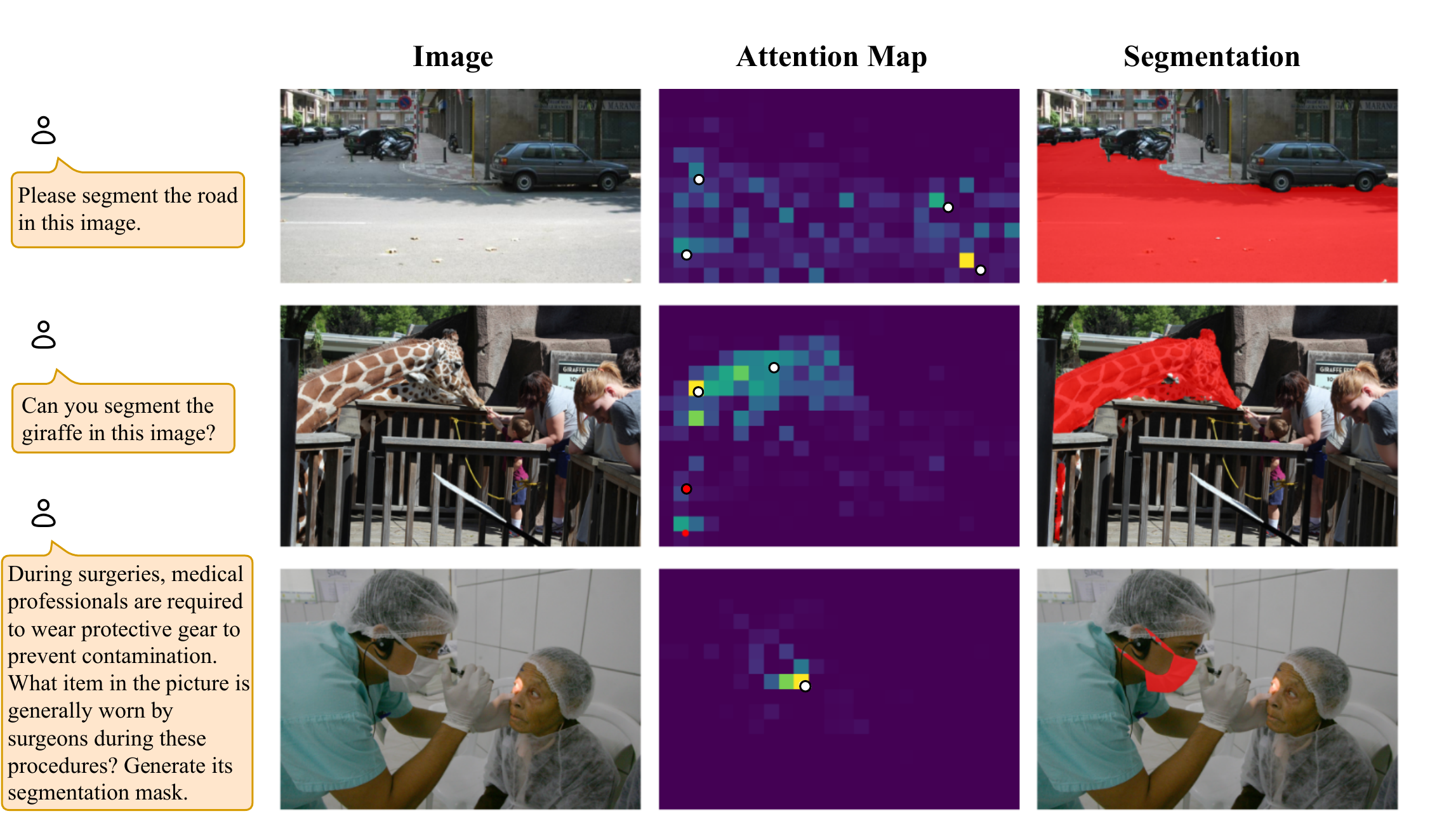}
\caption{{Showcases of \model.} Attention maps align well with ground-truth regions, where white points mark keypoints inside and red points mark those outside. These results illustrate how \model links semantics with segmentation}
    \label{fig:ssshow}
\end{figure*}

% 保留你原有的 Table 3 引用
% \input{tab/tab3}
\begin{table*}
    \centering
\caption{{Comprehensive comparison.}
\model achieves competitive segmentation performance under a substantially lighter training setting, while preserving the backbone’s general abilities.
Training memory values with underline mark inference overhead of the MLLM. 
\emph{Seg} denotes segmentation data (semantic, referring, reasoning),
\emph{FP-Seg} augments it with false-premise samples, and \emph{VQA} represents corpus from generative vision–language tasks. READ uses the largest training set, whereas \model relies solely on \emph{Seg}. {The bottom block reports \model on two additional backbones, Qwen2-VL-2B~\citep{wang2024qwen2vl} and LLaVA-OV-7B~\citep{li2024llava}, with the corresponding untouched base models as references (\cf \S\ref{sec:seg_eval})}}
    \label{table:efficiency}
        \raa{1.1}
    \setlength{\tabcolsep}{4pt}
    \resizebox{2\columnwidth}{!}{%
    \begin{tabular}{@{}l|l|r|ccc|r|rrrr@{}}
        \toprule
        \multirow{2}{*}{Method} 
        & \multirow{2}{*}{Backbone}
        & \multirow{2}{*}{\makecell{Training\\Mem (GB)$\downarrow$}} 
        & \multicolumn{3}{c|}{Training Data} 
        & \multirow{2}{*}{Seg $\uparrow$} 
        & \multirow{2}{*}{MME} & \multirow{2}{*}{MMBench} & \multirow{2}{*}{MMMU} & \multirow{2}{*}{MMStar}\\
        \cmidrule(lr){4-6}
        ~ & ~ & ~ & Seg & FP-Seg & VQA & ~ & ~ & ~ & ~ & ~ \\
        \midrule
        \texttt{random guess} & -- & -- & --  & --  & -- & -- & 1050.0 & 25.0 & 26.8 & 24.6\\
        LLaVA-1.5-7B & -- & -- &  &  & \checkmark & -- & 1808.4 & 66.5 & 35.7 & 33.1\\
        {Qwen2-VL-2B} & {--} & {--} &  &  & {\checkmark} & {--} & {1872.0} & {74.9} & {41.1} & {48.0}\\
        {LLaVA-OV-7B} & {--} & {--} &  &  & {\checkmark} & {--} & {1993.6} & {80.9} & {47.9} & {61.9}\\
        \midrule
        SESAME & LLaVA-1.5-7B & $30 \times 8$ &  & \checkmark & \checkmark & 30.4 & 1394.4 & 28.3 & 11.2 & 20.3\\
        LISA   & LLaVA-1.5-7B & $30 \times 8$ & \checkmark &  & \checkmark & 56.9 & 184.5 & 0 & 0 & 0\\
        READ   & SESAME       & $30 \times 8$ & \checkmark & \checkmark & \checkmark & {58.6} & 476.3 & 0 & 1.1 & 14.4 \\
        \model (Ours) & LLaVA-1.5-7B & \textbf{$\underline{16} + 10 \times 8$} & \checkmark &  &  & 57.3 & {1801.4} & {64.0} & {34.4} & {33.3}\\

        {\model (Ours)} & {Qwen2-VL-2B} & {$\underline{8}+10 \times 8$} & {\checkmark} &  &  & {54.3} & {1895.9} & {73.7} & {40.3} & {48.3}\\
        {\model (Ours)} & {LLaVA-OV-7B} & {$\underline{20}+10 \times 8$} & {\checkmark} &  &  & \textbf{{60.3}} & \textbf{{1987.7}} & \textbf{{80.6}} & \textbf{{48.1}} & \textbf{{61.1}}\\
        \bottomrule
    \end{tabular}
    }%
\end{table*}

\begin{table*}[htbp]
    \centering
\caption{{Comparisons on the ReasonSeg dataset.} The best performance is highlighted in \textbf{bold}, and the second best is \underline{underlined}. {The bottom block reports \model with two additional MLLM backbones (paired with SAM ViT-H).}
}
    \raa{1.0}
    \resizebox{2\columnwidth}{!}
    {
    \label{table:reason_seg}   
    \tabcolsep=0.3cm
    {
        % \begin{footnotesize}
        \begin{tabular}{@{}l|rr|rr|rr|rr@{}}
            \toprule
            
            \multirow{4}{*}{Method} & \multicolumn{2}{c|}{val} & \multicolumn{6}{c}{test} \\ 
            
            \specialrule{0em}{0pt}{1pt}
            \cmidrule{2-9}
            \specialrule{0em}{0pt}{1pt}

            ~ & \multicolumn{2}{c|}{overall} & \multicolumn{2}{c|}{short query} & \multicolumn{2}{c|}{long query} & \multicolumn{2}{c}{overall} \\

            \specialrule{0em}{0pt}{1pt}
            \cmidrule{2-9}
            \specialrule{0em}{0pt}{1pt}
            
            ~ & gIoU & cIoU & gIoU & cIoU & gIoU & cIoU & gIoU & cIoU \\ 
            
            \specialrule{0em}{0pt}{1pt}
            \hline
            \specialrule{0em}{0pt}{1pt}

            SESAME~\citep{wu2024see} & {34.8} & {39.1} & {28.3} & {27.6} & {31.6} & {32.7}& {30.5} & {30.4} \\
            
            LLaVA + OVSeg~\citep{liang2023open} & 38.2 & 23.5 & 24.2 & 18.7 & 44.6 & 37.1 & 39.7 & 31.8 \\

            LISA-7B~\citep{lai2024lisa} & 52.9 & 54.0 & 40.6 & 40.6 & 49.4 & 51.0 & 47.3 & 48.4 \\
            
            LISA-LLaVA-1.5-7B~\citep{lai2024lisa} & \underline{61.3} & 62.9 & \underline{48.3} & {46.3} & 57.9 & 59.7 & 55.6 & 56.9 \\

            READ-7B~\citep{read}  & 59.8 & \textbf{67.6} & \textbf{52.6} &\textbf{49.5}  & \textbf{60.4} & {61.0} &\textbf{58.5}  &\underline{58.6} 
            \\

            \specialrule{0em}{0pt}{1pt}
            \hline
            \specialrule{0em}{0pt}{1pt}
            \model-7B & \textbf{61.4} & {65.3} & 47.8 & 41.7 & \underline{59.3}&\underline{61.6} & \underline{56.5} & {57.3} \\ 
            \quad 14$^{th}$ layer $\to$ 30$^{th}$ layer & 45.7 & 43.6 &32.6 &35.7 &39.6 & 40.8 & 37.9  & 40.0\\
            \quad w/o keypoint description & 51.8 & 48.5 & 42.1 & 39.3 & 49.8 & 49.8 & 47.9  & 47.8\\ 
            \quad w/o global description & 56.0 & 61.9 & 44.0 & 40.3& 50.4& 49.4 & 48.8 & 47.9 \\  
            \quad w/o $\mathcal{L}_{\text{attn}}$ & 55.8 & 51.7 & 42.9 & 40.5 & 54.6 & 53.4 & 51.7 & 50.8\\
            \quad {w/ SAM ViT-B} & {55.9} & {59.8} & {45.3} & {44.8} & {52.5} & {54.9} & {50.7} & {52.8} \\
            \quad {w/ SAM ViT-L} & {60.8} & {64.4} & {45.6} & {42.5} & {54.1} & {59.1} & {52.1} & {55.4} \\
            \specialrule{0em}{0pt}{1pt}
            \hline
            \specialrule{0em}{0pt}{1pt}
            {\model\ (Qwen2-VL-2B)} & {57.7} & {63.8} & {48.0} & {\underline{47.8}} & {58.6} & {56.7} & {56.0} & {54.3} \\
            {\model\ (LLaVA-OV-7B)} & {59.6} & {\underline{67.3}} & {43.2} & {45.7} & {58.1} & \textbf{{64.3}} & {54.5} & \textbf{{60.3}} \\
            \bottomrule
        \end{tabular}
    }
    }
\end{table*}

\begin{table*}[htbp]
    \raa{1.}
    \centering
    \caption{{Comparison of SOTA referring segmentation (cIoU) on RefCOCO(+/g)}. }
    \label{table:refer_seg}   
    \resizebox{2\columnwidth}{!}{
    \tabcolsep=0.3cm
    {
        \begin{tabular}{@{} l | rrr | rrr | rr | r @{}}
            \toprule
            \multirow{3}*{Method} & \multicolumn{3}{c|}{RefCOCO} & \multicolumn{3}{c|}{RefCOCO+}  & \multicolumn{2}{c|}{RefCOCOg} & \multirow{3}*{Mean} \\ 
            \cmidrule{2-9}
            ~ & val & testA & testB & val & testA & testB & val(U) & test(U) & \\ 
            \midrule

            MCN~\citep{luo2020multi} & 62.4 & 64.2 & 59.7 & 50.6 & 55.0 & 44.7 & 49.2 & 49.4 & 54.4 \\

            VLT~\citep{ding2021vision} & 67.5 & 70.5 & 65.2 & 56.3 & 61.0 & 50.1 & 55.0 & 57.7 & 60.4 \\

            CRIS~\citep{wang2022cris} & 70.5 & 73.2 & 66.1 & 62.3 & 68.1 & 53.7 & 59.9 & 60.4 & 64.3 \\

            LAVT~\citep{yang2022lavt} & 72.7 & 75.8 & 68.8 & 62.1 & 68.4 & 55.1 & 61.2 & 62.1 & 66.5 \\
            
            ReLA~\citep{liu2023gres} & 73.8 & 76.5 & 70.2 & 66.0 & 71.0 & 57.7 & 65.0 & 66.0 & 68.3 \\
            
            X-Decoder~\citep{zou2023generalized} & -- & -- & -- & -- & -- & -- & 64.6 & -- & -- \\

            SEEM~\citep{zou2023segment} & -- & -- & -- & -- & -- & -- & 65.7 & -- & -- \\
            
            SESAME~\citep{wu2024see} & 74.7 & -- & -- & 64.9 & -- & -- & 66.1 & -- & -- \\
            
            LISA-LLaVA-1.5-7B~\citep{lai2024lisa} & 74.9 & {79.1} & {72.3} & 65.1 & 70.8 & 58.1 & 67.9 & {70.6} & 69.8 \\

            \model-7B & \textbf{76.5} & 78.3 & 71.4 & {66.1} & {71.7} & {58.3} & {69.4} & {70.6} & {70.3} \\
            {\model\ (Qwen2-VL-2B)} & {76.0} & {77.4} & \textbf{{73.5}} & {66.0} & {71.7} & {57.7} & {69.5} & {70.1} & {70.2} \\

            {\model\ (LLaVA-OV-7B)} & {75.5} & \textbf{{79.5}} & {72.2} & \textbf{{68.0}} & \textbf{{72.5}} & \textbf{{62.5}} & \textbf{{71.3}} & \textbf{{70.9}} & \textbf{{71.6}} \\

            \bottomrule
        \end{tabular}
    }
    }
\end{table*}

\noindent\textbf{Evaluations.} Our assessment proceeded from a \emph{comprehensive} perspective to a \emph{segmentation-specific} one. 
At the comprehensive level (\cf Table~\ref{table:efficiency}), \model excels in training efficiency while preserving general abilities {(all general-ability benchmarks are evaluated with their standard protocols)}.
At the segmentation level (\cf Table~\ref{table:reason_seg} and Table~\ref{table:refer_seg}), \model establishes competitive results on reasoning segmentation and RES, measured by gIoU (per-image IoU) and cIoU (dataset-level IoU).

\noindent\textbf{Baselines.}  
We directly compared against methods that require fine-tuning MLLMs for segmentation, including LISA~\citep{lai2024lisa}, SESAME~\citep{wu2024see}, and READ~\citep{read}. 
In addition, following LISA, we also included traditional baselines for RES task for further comparison on segmentation tasks, as reported in Table~\ref{table:refer_seg}.

\subsection{Comprehensive Evaluation}
We compared model backbones, training cost, training data, and resulting segmentation and general abilities (\cf Table~\ref{table:efficiency}). 
Training cost was measured under the DeepSpeed~\citep{rasley2020deepspeed} ZeRO-$2$ setting with $8$~GPUs{, under each method's official training configuration (\eg, a per-device batch size of 2 for the LoRA-based baselines; \cf \S\ref{sec:runtime} for the full protocol)}.
\emph{Seg} denotes the segmentation data (see \S\ref{trainingd}); 
\emph{FP-Seg} denotes an augmented version of \emph{Seg}, constructed using FP-RefCOCO(+/g)~\citep{wu2024see} and R-RefCOCO(+/g)~\citep{wu2024toward}.
\emph{VQA} is the instruction corpus from LLaVA. For segmentation evaluation, we reported cIoU on ReasonSeg. 
For general capability evaluation, we adopted MME~\citep{fu2023mme}, MMBench~\citep{liu2024mmbench}, MMMU~\citep{mmmu}, and MMStar~\citep{mmstar} benchmarks.

% \noindent\textbf{Training Efficiency.} 
% As shown in Table~\ref{table:efficiency}, \model is highly efficient. 
% Since the MLLM is used only for inference, gradients are not back-propagated through it, allowing distributed execution or even pre-caching. 
% As a result, the MLLM itself requires as little as $16$\,GB of memory. 
% Overall, \model reduces {the per-GPU training memory from 30\,GB to 10\,GB ($30\times8{=}240$\,GB $\to$ $10\times8+16{=}96$\,GB in total)} while still achieving the best comprehensive performance. {The full training-cost and inference-latency breakdown is deferred to \S\ref{sec:runtime} (Table~\ref{tab:runtime}).}

\noindent\textbf{Training Efficiency.}
{
As shown in Table~\ref{table:efficiency}, \model substantially reduces training memory by keeping the MLLM fully frozen and using it only for inference. Under the same 8-GPU setting, the training footprint is reduced from $30\times8$~GB for fine-tuning-based baselines to $10\times8+16$~GB for \model. A detailed breakdown of training memory, GPU-hours, and inference latency is provided in \S\ref{sec:runtime}.
}

\noindent\textbf{Avoiding the Multi-Objective Trade-off.} \model functions as a plug-and-play tool that the MLLM can invoke on demand (\eg, via chain-of-thought reasoning), without relying on special tokens or auxiliary objectives. It is trained exclusively on segmentation data and remains fully decoupled from the MLLM’s generative learning. By contrast, prior approaches entangle segmentation with generation through additional tokens and losses, which drastically compromises general-purpose ability. For instance, on MMBench, baselines such as LISA and READ collapse from $66.5\%$ to $0\%$. Crucially, this near-zero performance falls significantly below the random guess expectation ($ 25\%$ for 4-choice questions), indicating a catastrophic forgetting.

% \noindent\textbf{{Competitive} Segmentation with Preserved General Capabilities.}
% Our plug-and-play design endows \model with {segmentation ability on par with state-of-the-art fine-tuning-based methods} while retaining the general capabilities of the underlying MLLM.
% Compared to LISA, \model achieves {comparable} segmentation performance ($57.3$ \vs~$56.9$) without the collapse of general abilities (MMBench, MMMU, and MMStar all remain on par with LLaVA-1.5-7B, whereas LISA drops to zero).
% READ shows slightly better segmentation ($58.6${; a margin that falls within the sampling uncertainty of the benchmark, see \S\ref{sec:seg_eval}}) but benefits from a stronger backbone and larger training data, while still suffering from degraded generality.
% Overall, \model achieves {segmentation on par with the state of the art under the same lightweight training setting (\cf \S\ref{sec:discussion_sota} for a comparison with large-scale unified models)} while fully preserving general capabilities.
% {This preservation is not specific to LLaVA-1.5-7B: instantiating \model on two markedly different backbones, Qwen2-VL-2B and LLaVA-OV-7B (bottom block of Table~\ref{table:efficiency}), leaves every general-ability benchmark almost unchanged compared with the untouched base model (\eg, on Qwen2-VL-2B, MME 1895.9 \vs 1872.0 and MMStar 48.3 \vs 48.0), whereas the joint-training baselines degrade or collapse. How segmentation itself transfers across these backbones and across SAM decoder sizes is analyzed in \S\ref{sec:seg_eval}.}

\noindent\textbf{Competitive Segmentation with Preserved General Capabilities.}
{\model achieves segmentation performance comparable to fine-tuning-based methods while preserving the general capabilities of the underlying MLLM. For example, on ReasonSeg, \model achieves $57.3$ cIoU compared with $56.9$ for LISA, while MMBench, MMMU, and MMStar remain essentially unchanged from the frozen LLaVA-1.5-7B backbone, in contrast to the severe degradation of joint-training baselines. This preservation also holds for Qwen2-VL-2B and LLaVA-OV-7B, whose general-ability scores remain close to their untouched base models. Detailed segmentation results across datasets, MLLM backbones, and SAM decoder sizes are analyzed in \S\ref{sec:seg_eval}.}

\subsection{Segmentation Evaluation}
\label{sec:seg_eval}
We reported the segmentation performance of \model on both reasoning segmentation (Table~\ref{table:reason_seg}) and referring segmentation (Table~\ref{table:refer_seg}). Fig.~\ref{fig:ssshow} qualitatively illustrated the progression from attention maps to extracted keypoints and the final segmentation masks.

\noindent\textbf{Strong Performance on Both Reasoning and Referring Segmentation.} As shown in Table~\ref{table:reason_seg} and Table~\ref{table:refer_seg}, \model achieves strong performance on both reasoning segmentation (ReasonSeg) and referring segmentation (RefCOCO(+/g)).
On ReasonSeg, \model reaches $65.3$ cIoU on validation and $57.3$ on test, {comparable to} LISA ($62.9/56.9$) under the same LLaVA-1.5-7B backbone. On RefCOCO(+/g), \model achieves $70.3$, {on par with} LISA-LLaVA-1.5-7B ($69.8$) and markedly outperforming non-MLLM baselines such as LAVT ($66.5$) and CRIS ($64.3$).

\noindent{\textbf{Statistical Reliability.}
Small numerical margins on ReasonSeg should not be over-interpreted. Bootstrapping the ReasonSeg test set (10{,}000 resamples with replacement) yields 95\% confidence intervals of $[52.5, 62.7]$ for cIoU (reported: 57.3) and $[54.2, 59.4]$ for gIoU (reported: 56.5). The wider cIoU interval is expected: cIoU aggregates a cumulative intersection and union, so it is dominated by large-area objects and more sensitive to test-set composition at this scale, as noted by the ReasonSeg authors themselves~\citep{lai2024lisa}. We therefore treat margins within these intervals (\eg, the 1.3-point cIoU gap to READ) as comparable, applying this symmetrically and claiming no advantage for \model within such margins. The intervals do not, however, blur the effects our conclusions rest on: the contrast with SESAME (57.3 \vs 30.4 cIoU) and the $\sim$6.5-point ablation gain of the attention loss (Table~\ref{table:reason_seg}) both lie far outside any plausible sampling noise.}

\noindent\textbf{Improvement Room on Short Queries.}
ReasonSeg training set is highly imbalanced, as it was originally designed for training MLLMs with explanatory content. Since \model can only leverage the segmentation portion, which contains no short-query samples, performance on this category dataset remains limited. {Consistent with this view, Table~\ref{table:reason_seg} shows that all methods, not only \model, perform worse on the short split. Nevertheless, \model does not fail on this unseen query style: its overall ReasonSeg test performance remains comparable to LISA (57.3 \vs 56.9 cIoU), and it performs better on long queries, where the supervision is well covered (61.6 \vs 59.7 cIoU; 59.3 \vs 57.9 gIoU). These results suggest that \model retains a degree of query-style generalization, while the short split exposes a limitation of the current supervision coverage.}

\noindent{\textbf{Generalization across Backbones and Decoders.}
The attention--keypoint bridge is not tied to LLaVA-1.5-7B. On referring segmentation (Table~\ref{table:refer_seg}), the three backbones are nearly indistinguishable (mean cIoU 70.3/70.2/71.6); on ReasonSeg, segmentation tracks backbone capacity rather than any backbone-specific property: the stronger LLaVA-OV-7B lifts test cIoU to 60.3, so \model inherits backbone improvements for free, while the smaller Qwen2-VL-2B trails on reasoning (54.3) yet still matches the 7B models on referring expressions and stays competitive with the 7B LISA (val cIoU 63.8 \vs 62.9). Shrinking the SAM decoder degrades performance gracefully and monotonically (test cIoU: ViT-B 52.8 $\rightarrow$ ViT-L 55.4 $\rightarrow$ ViT-H 57.3; lower rows of Table~\ref{table:reason_seg}), mirroring the scaling of SAM itself~\citep{kirillov2023segment} and transferring across decoder sizes without modification. Together with the preserved general ability of Table~\ref{table:efficiency}, this confirms the bridge is a general, backbone- and decoder-agnostic mechanism rather than an artifact of LLaVA-1.5-7B.}

\subsection{Robustness and Ablation Analysis}
\label{sec:robustness}

To verify that \model is not reliant on fragile, handcrafted hyperparameter choices, we conducted a sensitivity analysis on the ReasonSeg test set.

\noindent\textbf{Component Analysis.} We evaluated the contributions of key components on the ReasonSeg dataset (\cf Table~\ref{table:reason_seg}). Removing the keypoint module or the attention loss lowers performance ($47.8$ and $50.8$), underscoring the importance of explicit spatial signals. Omitting the global description slightly affects validation performance but significantly harms test performance, highlighting its role in supporting generalization.

\noindent\textbf{Impact of Layer Selection.} 
A critical design choice is the selection of the intermediate layer. We hypothesize that shallow layers lack semantic understanding, while deep layers lose spatial granularity. 
To validate this, we evaluated performance using features extracted from different layers of the LLaVA-1.5-7B backbone. 
As shown in Table~\ref{tab:layer_ablation}, performance peaks around the middle layers (12--14). The choice of Layer 14 is thus empirically grounded in this semantic-spatial trade-off.

% 新增的 Ablation 表格，保持最基础的 table 环境，不改变排版风格
\begin{table}[h]
    \centering
    \caption{{Impact of layer selection} on ReasonSeg test set}
    \label{tab:layer_ablation}
    \begin{tabularx}{\linewidth}{@{}l *{6}{Y}@{}}
        \toprule
        Layer Index & 6 & 10 & 12 & \textbf{14 (Ours)} & 18 & 30 \\
        \midrule
        cIoU & 42.1 & 55.4 & 57.1 & \textbf{57.3} & 51.8 & 40.0 \\
        \bottomrule
    \end{tabularx}
\end{table}

\noindent\textbf{Sensitivity to Keypoint Number ($m$).}
We further analyzed the robustness of \model regarding the maximum number of sampled keypoints $m$.
As reported in Table~\ref{tab:k_ablation}, LENS performs best around $m=16$ and remains reasonably robust for smaller $m$, while too many keypoints can introduce noisy prompts and degrade performance.

\begin{table}[h]
    \centering
    \caption{{Sensitivity to the number of keypoints ($m$)}}
    \label{tab:k_ablation}
    \begin{tabularx}{\linewidth}{@{}l *{5}{Y}@{}}
        \toprule
        Num. Keypoints ($m$) & 4 & 8 & \textbf{16 (Ours)} & 24 & 32 \\
        \midrule
        cIoU & 55.6 & 57.0 & \textbf{57.3} & 51.1 & 52.5 \\
        \bottomrule
    \end{tabularx}
\end{table}

\begin{figure*}[t]
\centering
\includegraphics[width=0.96\linewidth]{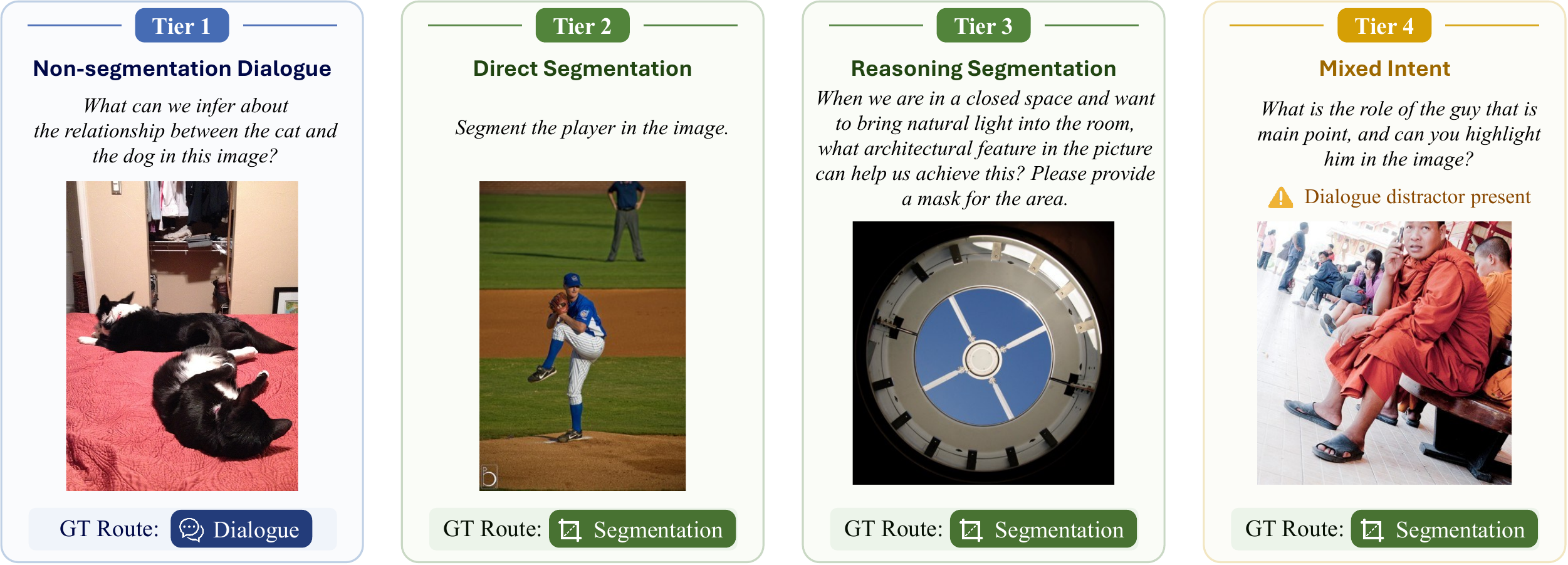}
\caption{{{Representative samples from our routing benchmark.} Each column shows one of the four difficulty tiers (non-segmentation dialogue, direct segmentation, reasoning segmentation, and mixed-intent), illustrating how the frozen MLLM router decides whether a query should be answered by dialogue or trigger the segmentation branch}}
\label{fig:routing_bench}
\end{figure*}

\noindent{\textbf{A Difficulty-Graded Routing Benchmark.}
Since \model operates within an agent framework where the frozen MLLM acts as a router, the reliability of this routing decision is critical. To evaluate it rigorously, we construct a difficulty-graded routing benchmark of 1{,}000 samples, curated by a Qwen3-VL-32B-assisted pipeline\footnote{\url{https://huggingface.co/Qwen/Qwen3-VL-32B-Instruct}}. Queries are sourced from the LLaVA instruction-tuning (VQA) corpus (dialogue), RefCOCO(+/g) (direct segmentation), and ReasonSeg (reasoning segmentation), with mixed-intent cases synthesized by combining them. To test robustness to prompt phrasing, the pipeline rewrites each segmentation phrase into diverse surface forms (\eg, ``\textit{segment \dots}'', ``\textit{generate the mask of \dots}'', ``\textit{please highlight \dots}''), so routing cannot rely on a fixed template. The frozen MLLM router must classify each query as \emph{Dialogue} or \emph{Segmentation} (the \emph{Follow-up} intent of Alg.~\ref{alg:agent_seg} is excluded, as it is resolved by the conversation memory state rather than by intent classification). The benchmark spans four difficulty tiers, one representative case per tier shown in Fig.~\ref{fig:routing_bench}:}
\begin{itemize}
\item {\textbf{Non-segmentation dialogue.} A general conversational/VQA query with no segmentation need; the correct route is \emph{Dialogue}, and any activation of the segmentation head counts as a false trigger. Example: ``\textit{What is the weather like in this photo?}''}
\item {\textbf{Direct segmentation.} A short instruction containing a literal segmentation command that names the target; the correct route is \emph{Segmentation}. Example: ``\textit{Segment the dog in this image.}''}
\item {\textbf{Reasoning segmentation.} A longer, descriptive query in which the segmentation instruction is only implied and the target must be inferred from the description; the correct route is still \emph{Segmentation}, but with much weaker surface cues. Example: ``\textit{Which object here is used to defend against predators? Generate its mask.}''}
\item {\textbf{Mixed-intent.} A single query blending several intents (\eg, a dialogue question and a segmentation request). We adopt the convention that an explicit segmentation request takes precedence, so the correct route is \emph{Segmentation}; treating such a query as dialogue-only counts as a missed trigger. Example: ``\textit{What breed is this dog, and can you also highlight it?}''}
\end{itemize}
{For each backbone (LLaVA-1.5-7B, Qwen2-VL-2B, LLaVA-OV-7B) we report per-tier routing accuracy and the overall false-/missed-trigger rates (Table~\ref{tab:routing_bench}). Pre-trained MLLMs handle this decision reliably: overall accuracy reaches 96.4\%--99.5\% without any routing-specific fine-tuning, supporting our claim that large-scale vision--language pre-training already provides strong semantic discrimination for the explicit or semantically recoverable cues that signal segmentation intent. The missed-trigger rate stays uniformly low ($\leq$1.7\%), so the router rarely overlooks genuine segmentation requests. The one visible weakness is the higher false-trigger rate of LLaVA-1.5-7B (9.2\%, \vs 0.0--0.4\% for the two newer backbones); as it drops sharply with backbone instruction-following strength, we attribute it to backbone capacity rather than the routing scheme. The benchmark construction pipeline is included in our released reproducibility package.}

\begin{tableorg}[t]
    \centering
    \caption{{{Routing evaluation on the difficulty-graded benchmark} (1{,}000 samples drawn from LLaVA VQA, RefCOCO(+/g), and ReasonSeg data, curated with a Qwen3-VL-32B-assisted pipeline). For each frozen MLLM router we report routing accuracy by tier and overall, together with the overall false-/missed-trigger rates (\%)}}
    \label{tab:routing_bench}
    \raa{1.1}
    \resizebox{\linewidth}{!}{%
    \begin{tabular}{@{}l|rrrr|r|rr@{}}
        \toprule
        \multirow{2}{*}{Router (backbone)} & \multicolumn{4}{c|}{Accuracy by tier} & \multirow{2}{*}{Overall} & \multirow{2}{*}{\makecell{False-\\trig.}} & \multirow{2}{*}{\makecell{Missed-\\trig.}} \\
        \cmidrule(lr){2-5}
        ~ & Non-seg & Direct & Reason. & Mixed & ~ & ~ & ~ \\
        \midrule
        LLaVA-1.5-7B & 90.8 & 99.2 & 98.0 & 97.6 & 96.4 & 9.2 & 1.7 \\
        Qwen2-VL-2B & 100.0 & 100.0 & 100.0 & 95.2 & 98.8 & 0.0 & 1.6 \\
        LLaVA-OV-7B & 99.6 & 100.0 & 100.0 & 98.4 & 99.5 & 0.4 & 0.5 \\
        \bottomrule
    \end{tabular}%
    }
\end{tableorg}

% 保持原有的图引用结构
\begin{figure*}[htbp] 
    \centering
    \begin{subfigure}[b]{\textwidth}
        \centering
        \includegraphics[width=0.88\linewidth]{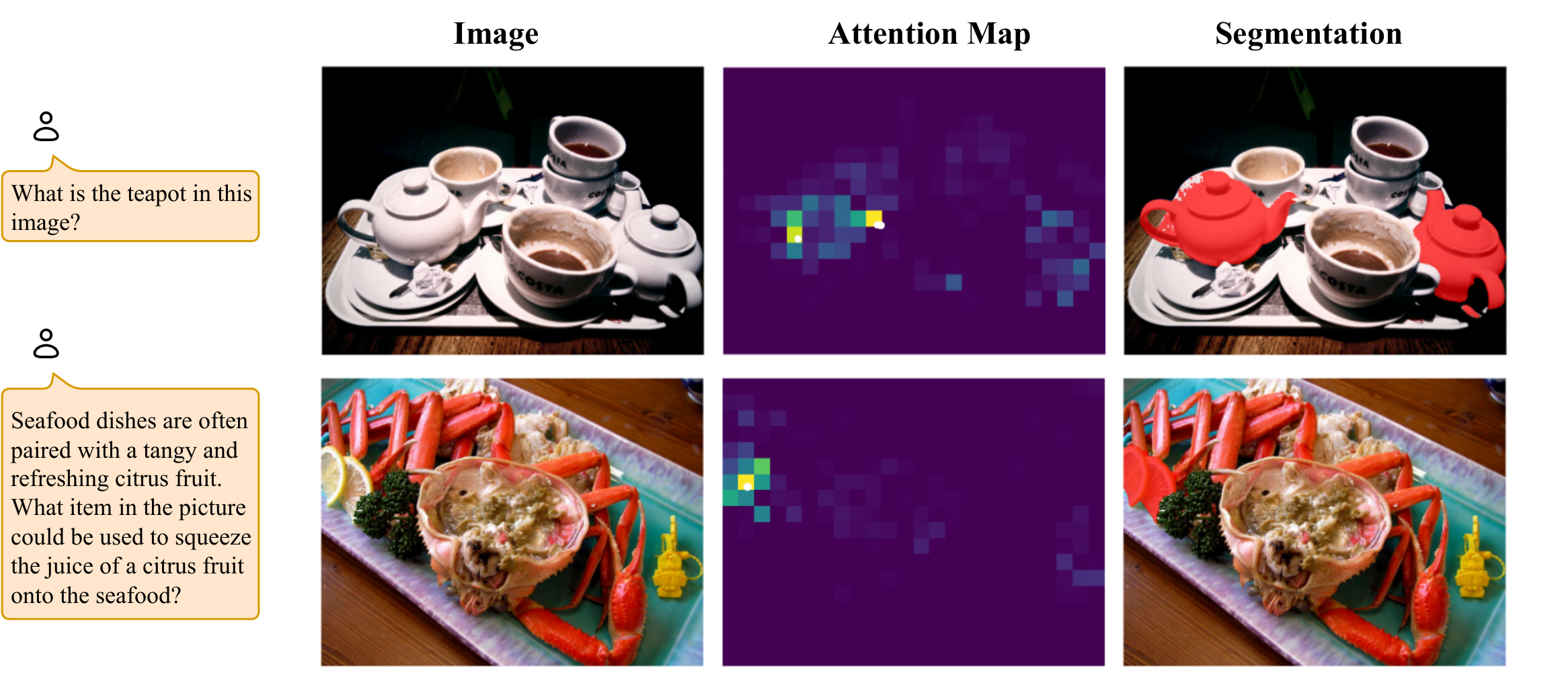}
        \caption{{Accurate Localization.} The model precisely focuses on the target object. White dots indicate valid keypoints}
        \label{fig:qual_success}
    \end{subfigure}
    \vspace{0.5em}
    \begin{subfigure}[b]{\textwidth}
        \centering
        \includegraphics[width=0.88\linewidth]{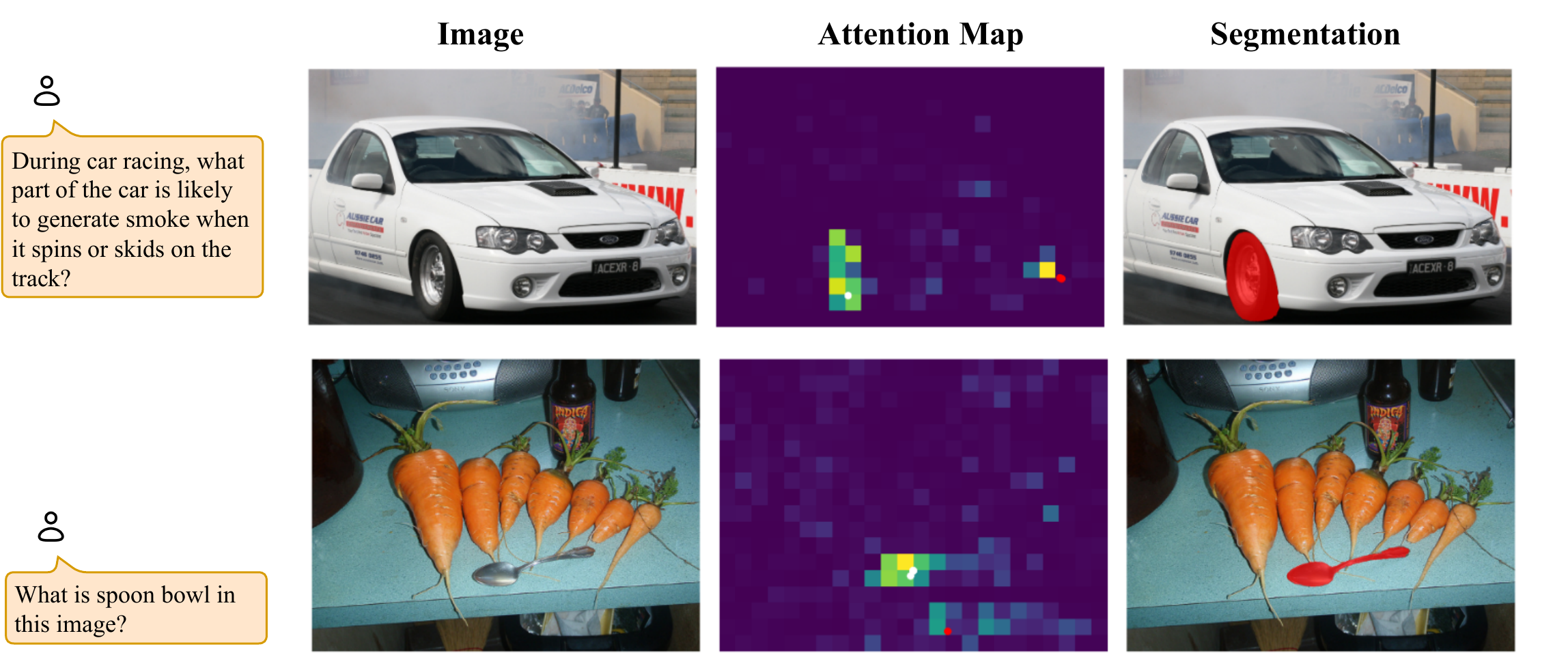}
        \caption{{Robust Description.} Even when attention maps activate on background regions (red dots), our descriptor module successfully rejects them, preserving the correct mask}
        \label{fig:qual_robust}
    \end{subfigure}
    \caption{{Qualitative results on the ReasonSeg dataset.} 
    Visualizing the progression from input instruction to attention maps, keypoint extraction, and final segmentation}
    \label{fig:qualitative_analysis}
\end{figure*}

\begin{figure*}[t]
    \centering
    \includegraphics[width=0.88\textwidth]{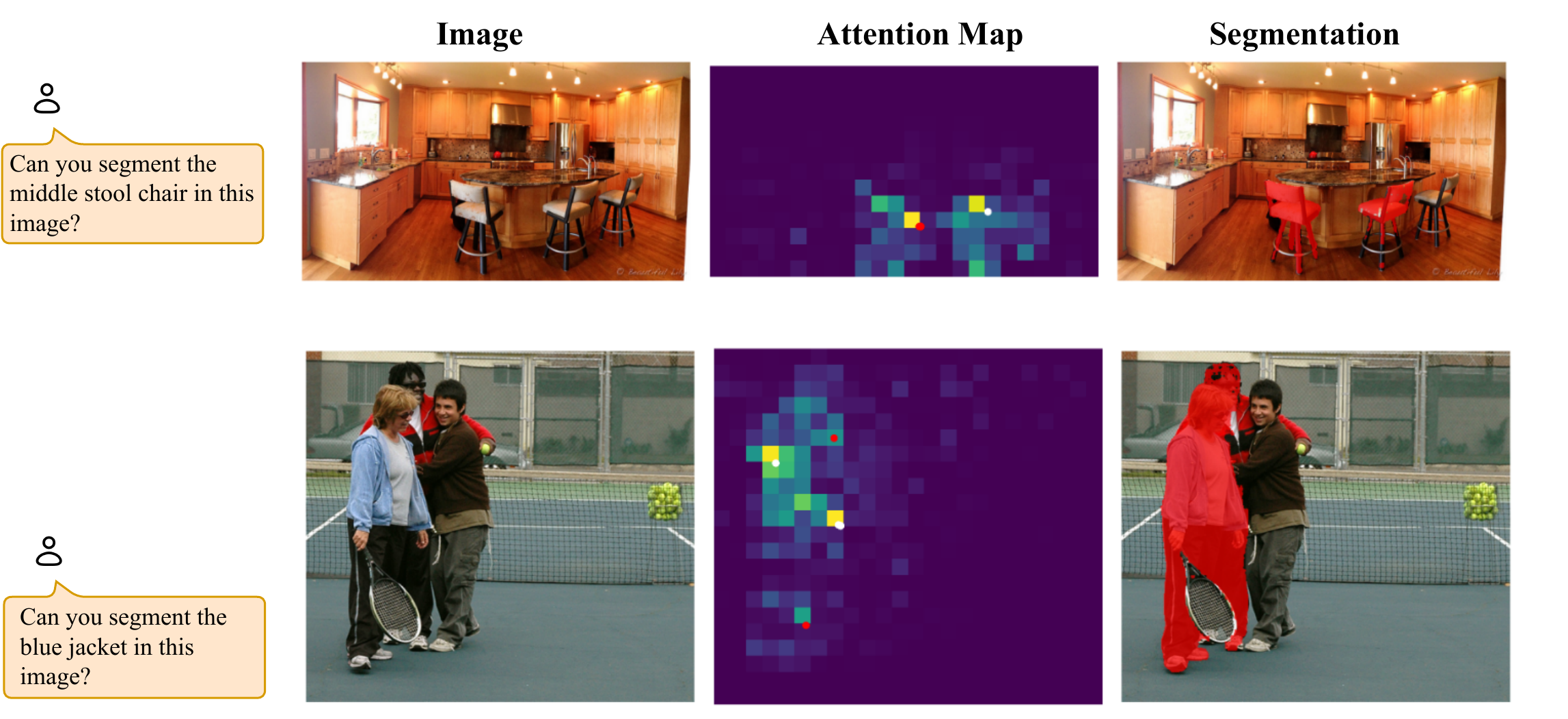}
    \caption{{Failure cases.} The primary limitation of \model is its dependency on the frozen backbone}
    \label{fig:failure_cases}
\end{figure*}

\subsection{Runtime and Computational Complexity}
\label{sec:runtime}

{We compare training cost and inference efficiency against LISA, SESAME, and READ under a common protocol (the same LLaVA-1.5-7B architecture, with READ initialized from the SESAME checkpoint as in Table~\ref{table:efficiency}; the same SAM encoder; $8\times$A800 GPUs, bf16). Table~\ref{tab:runtime} reports, under each method's official configuration, peak training memory across 8 GPUs, trainable parameters, training samples processed, total GPU-hours, and per-image latency. The key distinction is architectural: \model trains an \emph{external} grounding head on top of a fully frozen MLLM, whereas the baselines inject LoRA modules \emph{inside} the MLLM.}

\noindent{\textbf{Memory and Training Compute.}
Notably, \model's external head has \emph{more} trainable parameters than the LoRA baselines (668M \vs $\sim$305M); the efficiency gain comes from \emph{where} those parameters sit, not from their number. LoRA modules live inside the 7B backbone, so every step back-propagates through the full MLLM graph and stores its activations. In contrast, \model keeps the MLLM in a no-grad forward mode: the frozen backbone is served as a single shared instance by an inference engine while only the external head is optimized. This avoids storing full-backbone activations and gradients, reducing peak memory from $30\times8$~GB to $10\times8+16$~GB (a 10~GB head per GPU plus one 16~GB frozen backbone shared by all 8 GPUs).}

\noindent{\textbf{Training Samples and Time.}
\model converges after only 256k training samples (10 epochs $\times$ 100 steps $\times$ an effective batch size of $8\times32=256k$), about $3\times$ fewer than LISA and READ ($\sim$800k). SESAME uses a comparable count (250k), but its augmented, longer instructions inflate the per-step cost: SESAME spends $\sim$30 GPU-hours \vs $\sim$16 for \model. Together with the absence of any backbone backward pass, \model attains the lowest total training cost.}

\noindent{\textbf{Latency and Complexity.}
Once routed to the segmentation branch, \model does not need to autoregressively decode a \texttt{[SEG]} token: it reads the start-of-answer feature from a single MLLM prefill and invokes the grounding/segmentation head directly. The baselines instead decode a special \texttt{[SEG]} token before the mask decoder, costing one prefill plus $T$ sequential decoding steps that each traverse all 7B layers, whereas \model costs the prefill only plus a head whose FLOPs are negligible. Eliminating the $T$-step stage yields the 6--9$\times$ per-image speedup (182~ms \vs 1104--1631~ms). In sum, \model attains comparable segmentation quality with lower training memory, fewer GPU-hours, and a fraction of the inference latency.}

\begin{table*}[t]
    \centering
    \caption{{{Training cost and inference efficiency.} All methods build on the LLaVA-1.5-7B architecture (READ is initialized from the SESAME checkpoint) and use the same SAM encoder under their official training configurations ($8\times$A800, bf16). \emph{Peak Mem.}: total peak training memory across 8 GPUs. \emph{Train.\ Params}: trainable parameters and where they reside. \emph{Train.\ Samples}: total training samples processed until convergence under each method's official recipe (epochs $\times$ steps $\times$ effective batch size; for \model, $10\times100\times(8\times32)=256$k). \emph{Latency}: per-image inference latency with batch size 1, bf16, on one A800}}
    \label{tab:runtime}
    \raa{1.1}
    \setlength{\tabcolsep}{4pt}
    \resizebox{2\columnwidth}{!}{%
    \begin{tabular}{@{}l|r|r|r|r|r@{}}
        \toprule
        Method & Peak Mem.\ (GB, 8 GPUs) & Train.\ Params (M) & Train.\ Samples & Train.\ Time (GPU-h) & Latency (ms) $\downarrow$ \\
        \midrule
        SESAME & $30\times8$ & $\sim$305 (LoRA, in MLLM) & 250k & $\sim$30 & 1340 \\
        LISA & $30\times8$ & $\sim$305 (LoRA, in MLLM) & 800k & $\sim$53 & 1104 \\
        READ & $30\times8$ & $\sim$305 (LoRA, in MLLM) & 800k & $\sim$56 & 1631 \\
        \model\ (Ours) & $\mathbf{10\times8+16}$ & 668 (external head) & 256k & $\sim$\textbf{16} & \textbf{182} \\
        \bottomrule
    \end{tabular}%
    }
\end{table*}

\subsection{Efficiency and SOTA}
\label{sec:discussion_sota}

Recent advancements in unified multimodal models, such as GLaMM~\citep{rasheed2024glamm} and SegLLM~\citep{wang2025segllm}, have pushed the boundaries of segmentation performance.
\textbf{Quantitatively}, on the RES benchmarks (RefCOCO/+/g splits), GLaMM and SegLLM achieve impressive mean cIoU scores of {75.6} and {73.6}, respectively. 
Compared to \model's average of {70.3}, there exists a performance gap. 
However, this gap must be contextualized within the trade-offs of {data scale} and {training paradigm}.

\noindent\textbf{Data Efficiency.} 
State-of-the-art methods typically rely on massive composite datasets (e.g., GLaMM uses millions of samples). 
In contrast, \model follows the strictly lightweight setting of LISA~\citep{lai2024lisa}, training solely on the segmentation subset (less than 10\% of the data used by large-scale baselines). 
Despite this data disparity, \model achieves competitive performance, highlighting its high data efficiency.

\noindent\textbf{Comparison Logic.} 
Most SOTA methods require full parameter fine-tuning. 
\model offers a distinct advantage by freezing the backbone and training only a lightweight head. 
While \model may not currently hold the absolute SOTA on leaderboards populated by heavy models, it establishes a pareto-optimal frontier for {efficient, safe, and trustworthy} extensions of MLLMs.

\subsection{Qualitative Analysis}
\label{sec:qualitative_analysis}

To provide deeper insights into the internal mechanisms of \model, we visualize the intermediate outputs, specifically the attention heatmaps, the extracted keypoints, and the final segmentation masks, across diverse scenarios in Fig.~\ref{fig:qualitative_analysis}.

\noindent\textbf{Interpreting Visual Grounding.} 
As observed in the first two rows of Fig.~\ref{fig:qualitative_analysis}, the MLLM's internal attention maps ($A_c^1$) successfully localize the semantically relevant regions. 
\model leverages these intrinsic cues to sample candidate keypoints (marked as dots). 
Crucially, the sub-pixel refinement strategy allows these keypoints to align precisely with the object boundaries.

\noindent\textbf{Robustness via Description.} 
The third and fourth rows of Fig.~\ref{fig:qualitative_analysis} demonstrate the critical role of our \textit{Keypoint Description} module. 
In these cases, the attention maps exhibit high responses on non-target regions, leading to the extraction of false-positive keypoints (marked in red). 
However, our descriptor module correctly identifies these keypoints as negative samples. This ``Select-and-Verify" mechanism effectively decouples the localization recall from the segmentation precision.

\section{Discussion and Limitations}
\label{sec:limitations}

\begin{figure}[t]
    \centering
    \includegraphics[width=\linewidth]{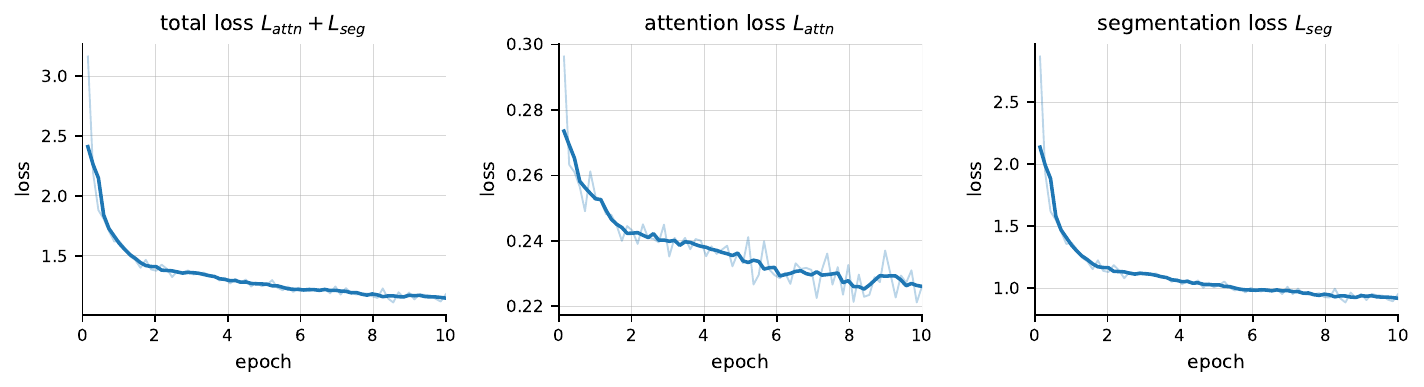}
    \caption{{{Training-loss curves of \model\ (LLaVA-OV-7B backbone).} The total training loss and its two components, the attention loss $\mathcal{L}_{\text{attn}}$ and the segmentation loss $\mathcal{L}_{\text{seg}}$, decrease smoothly and plateau over the 10 training epochs, indicating stable optimization without divergence or oscillation. Light curves show raw logged values; solid curves apply a centered moving average}}
    \label{fig:loss_curves}
\end{figure}

\noindent{Two further analyses support \model's practical value. In a blind study on 100 ReasonSeg/RefCOCO cases (three computer-vision-background raters; the masks of \model, LISA, and the ground truth shown anonymized and in randomized order), \model's masks are rated on par with LISA (3.9 \vs 3.7 on a 5-point Likert scale), while its exposed positive/negative keypoints are rated interpretable (4.1/5)---a verification handle for users and downstream agents that an opaque \texttt{[SEG]} token cannot offer (Table~\ref{tab:human_eval}). Training is likewise stable: under standard BCE$+$Dice and BCE objectives with AdamW, the total loss and its two components decrease smoothly and plateau over the 10 epochs without divergence or oscillation (Fig.~\ref{fig:loss_curves}).}

\begin{table}[t]
    \centering
    \caption{{{Human evaluation on 100 segmentation cases} (ReasonSeg/RefCOCO; $N{=}3$ raters; 5-point Likert means $\uparrow$). Keypoint interpretability, which we use to gauge agent usability, applies only to methods exposing intermediate keypoints}}
    \label{tab:human_eval}
    \begin{tabularx}{\linewidth}{@{}l Y Y@{}}
        \toprule
        Method & Mask quality $\uparrow$ & Keypoint interp.\ $\uparrow$ \\
        \midrule
        GT (reference) & 4.9 & -- \\
        LISA & 3.7 & -- \\
        \model\ (Ours) & 3.9 & 4.1 \\
        \bottomrule
    \end{tabularx}
\end{table}

While \model offers a highly efficient and effective solution for unified vision modeling, it inherits certain limitations stemming from its reliance on a completely frozen backbone.

\noindent\textbf{Dependency on Backbone Attention.} 
The most significant limitation arises from the ``upper bound" set by the MLLM's intrinsic visual perception. 
Since \model does not finetune the MLLM, it relies entirely on the attention maps produced during inference. 
As illustrated in Fig.~\ref{fig:failure_cases}, if the MLLM hallucinates or fails to attend to the target object due to severe occlusion, the segmentation head receives sparse or misleading spatial cues.

\noindent{\textbf{Requirement of Intermediate-Feature Access.}
As noted in \S\ref{sec:head}, \model targets the \emph{frozen-but-accessible} setting: the MLLM is never updated, but its intermediate attention maps and features must be accessible for deriving keypoints. \model is thus not applicable to strictly black-box, API-only models that expose text outputs alone. This does not affect our main claim of preserving the backbone and avoiding segmentation-specific fine-tuning.}

\noindent{\textbf{Granularity on Tiny Objects and Coarse-to-Fine Mitigation.}
{
Another limitation arises from the spatial resolution of the attention map used for keypoint extraction. Even with sub-pixel refinement, extremely small or dense objects remain challenging when their footprint occupies only a small fraction of the coarse attention grid ($24\times24$ for LLaVA-1.5-7B). We mitigate this issue using a simple training-free coarse-to-fine strategy that keeps the MLLM entirely frozen. Standard \model first produces a coarse target estimate on the full image; we then enlarge the bounding box of the predicted mask to preserve context (or center the crop at the attention maximum when the prediction is empty), resize the crop to the backbone input resolution, and re-process it using the same frozen pipeline. This allocates more visual tokens to the target region and therefore provides finer spatial cues for keypoint localization, without introducing additional parameters or supervision.

To quantify this effect, we construct a small-object subset of the ReasonSeg test split by selecting short queries whose target-mask area is below 10\% of the image, yielding 140 challenging cases. As shown in Table~\ref{tab:small_obj}, the refinement improves gIoU from 40.8 to 43.9 on LLaVA-1.5-7B and from 37.0 to 44.3 on LLaVA-OV-7B. These consistent gains across two different frozen backbones indicate that insufficient spatial resolution is an important source of small-object errors. Cropping around the coarse target estimate and reprocessing the crop at the same backbone input resolution allows the target to occupy more visual tokens, thereby reducing the effective spatial quantization of the attention map and improving keypoint localization. Nevertheless, the refinement cannot correct cases where the frozen MLLM attends to an incorrect region in the first place, such as attention hallucination or heavily occluded targets.
}

}

\begin{table}[t]
    \centering
    % \caption{{{Mitigating the small-object limitation (training-free, frozen backbone).} Per-instance gIoU on a small-object subset of the ReasonSeg test split (short queries with target-mask area $<10\%$; 140 cases). The mitigation is a single coarse-to-fine pass: a first full-image prediction localizes the target, the same frozen pipeline then re-segments an enlarged crop around it, and the resulting mask is pasted back; no parameters are added and no fine-tuning is performed}}
    \caption{{{Training-free coarse-to-fine mitigation for small objects.} Per-instance gIoU on the ReasonSeg small-object subset (short queries with target-mask area $<10\%$; 140 cases). The same frozen pipeline is evaluated on the full image and with coarse-to-fine refinement; no additional parameters or fine-tuning are introduced}}
    \label{tab:small_obj}
    \begin{tabularx}{\linewidth}{@{}l l Y@{}}
        \toprule
        Backbone & Setting & gIoU \\
        \midrule
        \multirow{2}{*}{LLaVA-1.5-7B} & full image (single pass) & 40.8 \\
        ~ & $+$ coarse-to-fine & 43.9 (+3.1) \\
        \midrule
        \multirow{2}{*}{LLaVA-OV-7B} & full image (single pass) & 37.0 \\
        ~ & $+$ coarse-to-fine & 44.3 (+7.3) \\
        \bottomrule
    \end{tabularx}
\end{table}

{\noindent{\textbf{Potential Extension to Video.}
Although \model currently derives keypoint prompts from single-image MLLM attention, a natural extension is to apply \model on selected video frames to generate semantic keypoint prompts and use video-capable segmentation models such as SAM 2~\citep{ravi2024sam} to propagate the resulting masks across frames.
}
}

\noindent{\textbf{Broader Impact.}
Stronger visual grounding and segmentation carry dual-use risks, including potential misuse for surveillance. \model is intended for research and assistive visual understanding, not for identifying or tracking individuals. Because the MLLM backbone is frozen, \model does not alter its original safety guardrails or refusal behaviors; we nonetheless emphasize that deployment should follow privacy-preserving policies and avoid sensitive surveillance scenarios.}

\section{Conclusion}

This work establishes \model (\textbf{L}everaging k\textbf{E}ypoi\textbf{N}ts for MLLMs' \textbf{S}egmentation) as a plug-and-play architecture that brings segmentation into MLLMs without compromising their general-purpose abilities. By freezing the entire MLLM and introducing a lightweight head that leverages the model’s own spatial cues as keypoints, \model bypasses the objective conflict that hampers prior fine-tuning-based approaches. Our experiments demonstrate that \model achieves {segmentation performance on par with state-of-the-art fine-tuning-based methods under the same lightweight training setting} while preserving the MLLM’s broad capabilities and cutting training costs by a large margin. These results highlight \model as an efficient and scalable paradigm for extending MLLMs, marking a step toward trustworthy and unified vision models. By reconciling task-specific precision with general-purpose integrity, \model paves the way for reliable agents capable of navigating the full spectrum of open-world visual tasks.

\section*{Statements and Declarations}

\begin{itemize}
\item \textbf{Funding}: This work was supported by National Natural Science Foundation of China (NSFC) Young Scientists Fund Category B (62522216), Hong Kong SAR Research Grants Council (RGC) General Research Fund (16219025), National Natural Science Foundation of China (NSFC) Young Scientists Fund Category C (62402408), and Hong Kong SAR Research Grants Council (RGC) Early Career Scheme (26208924).

\item \textbf{Competing Interests}: The authors have no relevant financial or non-financial interests to disclose.

\item \textbf{Data Availability}: The datasets analyzed during the current study are available in their respective public repositories: COCO-Stuff (\url{https://github.com/nightrome/cocostuff}), ADE20K (\url{http://sceneparsing.csail.mit.edu/}), RefCOCO/+/g (\url{https://github.com/lichengunc/refer}), and ReasonSeg (\url{https://github.com/dvlab-research/LISA}).

\item \textbf{Code Availability}: {The code and supporting materials, including training and evaluation scripts, experiment configurations, and the routing-benchmark construction pipeline, are publicly available at \url{https://github.com/HKUST-LongGroup/LENS}.}

\item \textbf{Consent for Publication}: Not applicable.

\item \textbf{Author Contribution}: All authors contributed to the study conception and design. Material preparation, data collection, and analysis were performed by Jiazhen Liu. The first draft of the manuscript was written by Jiazhen Liu. Long Chen provided supervision and funding support. All authors read and approved the final manuscript.
\end{itemize}

\bibliographystyle{sn-basic}
\bibliography{sn-bibliography}% common bib file

\end{document}